\documentclass[10pt,journal,final]{IEEEtran}
\usepackage[latin9]{inputenc}
\usepackage{url}
\usepackage{amsmath}
\usepackage{amssymb}
\usepackage{graphicx}

\makeatletter
\usepackage{textcomp}
\usepackage{amsfonts}
\usepackage{comment}
\ifCLASSINFOpdf
\else
\fi
\makeatother

\begin{document}
\title{Practical Distributed Control for VTOL UAVs to Pass a Virtual Tube}
\author{Quan~Quan,~ Rao Fu, Mengxin Li, Donghui Wei, Yan Gao and Kai-Yuan
Cai \thanks{Q. Quan, R. Fu, M. Li, Y. Gao and K-Y. Cai are with the School of
Automation Science and Electrical Engineering, Beihang University,
Beijing 100191, China (e-mail: qq\_buaa@buaa.edu.cn; buaafurao@buaa.edu.cn;
lmxin@buaa.edu.cn; buaa\_gaoyan@buaa.edu.cn; kycai@buaa.edu.cn).}\thanks{D. Wei is with the Beijing Electro-Mechanical Engineering Institute,
Beijing 100074, China (e-mail: weidonghui2652@sina.com).}}
\maketitle
\begin{abstract}
Unmanned Aerial Vehicles (UAVs) are now becoming increasingly accessible
to amateur and commercial users alike. An air traffic management (ATM)
system is needed to help ensure that this newest entrant into the
skies does not collide with others. In an ATM, airspace can be composed
of \emph{airways, intersections }and\emph{ nodes. }In this paper,
for simplicity, distributed coordinating the motions of Vertical TakeOff
and Landing (VTOL) UAVs to pass an \emph{airway }is focused. This
is formulated as a \emph{virtual tube passing problem}, which includes
passing a virtual tube, inter-agent collision avoidance and keeping
within the virtual tube. Lyapunov-like functions are designed elaborately,
and formal analysis based on \textit{invariant set theorem} is made
to show that all UAVs can pass the virtual tube without getting trapped,
avoid collision and keep within the virtual tube. What is more, by
the proposed distributed control, a VTOL UAV can keep away from another
VTOL UAV or return back to the virtual tube as soon as possible, once
it enters into the safety area of another or has a collision with
the virtual tube during it is passing the virtual tube. Simulations
and experiments are carried out to show the effectiveness of the proposed
method and the comparison with other methods. 
\end{abstract}

\begin{IEEEkeywords}
Distributed control, swarm, UAVs, air traffic, virtual tube. 
\end{IEEEkeywords}

\section{Introduction}

\begin{figure}[h]
\begin{centering}
\includegraphics[scale=0.65]{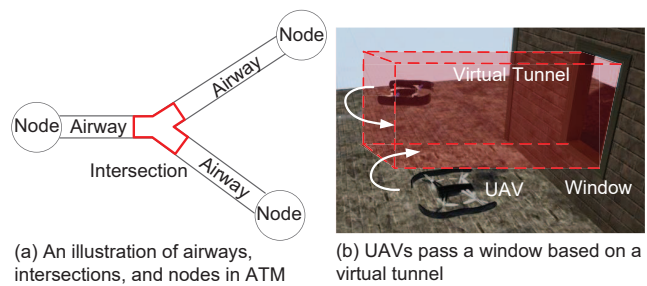} 
\par\end{centering}
\caption{Practical application scenarios of virtual tube passing problem.}
\label{Airspacestructure} 
\end{figure}

Airspace is utilized today by far lesser aircraft than it can accommodate,
especially low altitude airspace. There are more and more applications
for UAVs in low altitude airspace, ranging from the on-demand package
delivery to traffic and wildlife surveillance, inspection of infrastructure,
search and rescue, agriculture, and cinematography. Moreover, since
UAVs are usually small owing to portability requirements, it is often
necessary to deploy a team of UAVs to accomplish certain missions.
All these applications share a common need for both navigation and
airspace management. One good starting point is NASA's Unmanned Aerial
System Traffic Management (UTM) project, which organized a symposium
to begin preparations of a solution for low altitude traffic management
to be proposed to the Federal Aviation Administration. What is more,
air traffic for UAVs is attracted more and more research \cite{IoD(2016)},\cite{Devasia(2016)}.
Traditionally, the main role of air traffic management (ATM) is to
keep a prescribed separation among all aircraft by using centralized
control. However, it is infeasible for increasing UAVs because the
traditional control method lacks scalability. In order to address
such a problem, free flight is a developing air traffic control method
that uses no centralized control. Instead, parts of airspace are reserved
dynamically and automatically in a distributed way using computer
communication to ensure the required separation among aircraft. This
new system may be implemented into the U.S. air traffic control system
in the next decade. Airspace may be allocated temporarily by an ATM
for a special task within a given time interval. In this airspace,
these aircraft have to be managed so that they can complete their
tasks meanwhile avoiding collision. In \cite{IoD(2016)}, the airspace
is structured similarly to the road network as shown in Figure \ref{Airspacestructure}(a).
Aircraft are only allowed inside the following three: \emph{airways}
playing a similar role to roads or virtual tubes, \emph{intersections}
formed by at least two airways, and \emph{nodes} which are the points
of interest reachable through an alternating sequence of airways and
intersections. 

In this paper, for simplicity, coordinating the motions of VTOL UAVs
to pass an \emph{airway }is considered, which can be taken as a virtual
tube or corridor in the air. Concretely, the main problem is to coordinate
the motions of VTOL UAVs include passing a virtual tube, inter-agent
conflict (coming within the minimum allowed distance between each
other, not to be confused with a \emph{collision}) avoidance and keeping
within the virtual tube, which is called the \emph{virtual tube passing
problem }here, which is very common in practice. For example, virtual
tubes can be paths connecting two places, designed to bypass areas
having the dense population or to be covered by wireless mobile networks
(4G or 5G). virtual tubes can also be gates, corridors or windows,
because they can be viewed as virtual virtual tubes, as shown in Figure
\ref{Airspacestructure}(b). Such problems of coordination of multiple
agents have been addressed partly using different approaches, various
stability criteria and numerous control techniques \cite{Parker(2009)},\cite{Ren(2011)},\cite{Antonelli(2013)},\cite{Yan(2013)},\cite{Hoy(2015)},\cite{Oh(2015)},\cite{Survey}.
A commonly-used method, namely the dynamic region-following formation
control, is to organize multiple agents as a group inside a region
and then move the group to pass virtual tubes, where the size of the
region can vary according to the virtual tubes \cite{Hou(2009)},\cite{Chen(2018)},\cite{Dutta(2018)},\cite{Wang(2007)}.
However, the formation control is not very suited for the air traffic
control problem considered. First, each UAV has its own task, while
the formation control means that one has to wait for other ones. Second,
higher-level coordination should be made to decide which ones should
be in one group. What is more, the number of UAVs in airspace is varying
dynamically, which increases the design difficulty of the higher-level
coordination. Another way is to plan the trajectories for UAVs \cite{Liu(2019)},\cite{Capt},\cite{Ingersoll(2016)}.
However, planning often depends on global information and may have
to be updated due to uncertainties in practice, which brings more
complex calculations.

According to the consideration above, we propose distributed control
for VTOL UAV swarm, each one having the same control protocol. Distributed
control will not use the global information so that the computation
only depends on the number of local UAVs \cite{Connectedness},\cite{Liao(2017)},\cite{Distributed},\cite{Zhu(2015)}.
This framework is applicable to dense air traffic. By the proposed
protocol, every UAV can pass a virtual tube freely not in formation,
meanwhile avoiding conflict with each other and keeping within the
virtual tube once it enters into the virtual tube. During the process,
some UAVs with high speed will overtake slow ones. The idea used is
simialr to artificial potential field methods because of its ease-of-use,
where designed barrier functions \cite{Safety} are taken as artificial
potential functions. The distributed control laws use the negative
gradient of mixing of attractive and repulsive potential functions
to produce vector fields that ensure the passing and conflict avoidance,
respectively. However, it is not easy to use such an idea, with two
reasons in the following. 
\begin{itemize}
\item The guidance strategy for each UAV has to design. An easy method is
to set a chain of waypoints for each UAV. However, UAVs may get trapped
when using this method. Namely, they have not arrived at their corresponding
waypoints, but velocities are zero. Consequently, in order to avoid
trap, a higher-level decision should be made to set these waypoints.
As indicated by \cite{Hernandez(2011)}, the complexity of the calculation
of undesired equilibria remains an open problem. An example is proposed
in \cite{Ingersoll(2016)}, modelling the virtual tube passing problem
as an objective optimization problem, which can get the optimal path
to minimize the length, time or energy by designing suitable optimization-based
algorithm and objective functions. This algorithm works well for offline
path-planning. However, if the obstacles to avoid are dynamic, the
optimization-based algorithm will consume a lot of time to update
global information and the corresponding constraints during the online
path-planning process, which is not suitable for dense air traffic
because of lacking real-time of control. Compared to the optimal solution
of targets, safety, real-time and reliability are more necessary in
practice. 
\item Besides this problem, the second problem is also encountered in practice
especially for UAVs outdoor. The conflict of two agents is often defined
in control strategies that their distance is less than a safety distance.
The area is called \emph{safety area} of an agent if the distance
to the agent is less than the safety distance. However, a conflict
will happen in practice even if conflict avoidance is proved formally
because some assumptions will be violated in practice. For example,
a UAV may enter into the safety area of another due to an unpredictable
communication delay. On the other hand, most likely, two UAVs may
not have a real collision in physics because the safety distance is
often set large by considering various uncertainties, such as estimate
error, communication delay, and control delay. This is a big difference
from some indoor robots with a highly accurate position estimation
and control. In most literature, if their distance is less than a
safety distance, then their control schemes either do not work or
even push the agent towards the center of the safety area rather than
leaving the safety area. For example, some studies have used the following
barrier function terms for collision avoidance, such as $1\mathord{\left/\left(\left\Vert {\bf p}_{i}-{\bf p}_{j}\right\Vert ^{2}-R\right)\right.}$\cite[p. 323]{Quan(2017)}
or $\ln\left(\left\Vert \mathbf{p}_{i}-\mathbf{p}_{j}\right\Vert -R\right)$
\cite{Panagou(2016)}, where $\mathbf{p}_{i}$, $\mathbf{p}_{j}$
are two UAVs'{} positions, and $R>0$ is the separation distance.
The principle is to design a controller to make the barrier function
terms bounded so that $\left\Vert \mathbf{p}_{i}-\mathbf{p}_{j}\right\Vert ^{2}>R$
if $\left\Vert \mathbf{p}_{i}\left(0\right)-\mathbf{p}_{j}\left(0\right)\right\Vert ^{2}>R$.
Otherwise, $\left\Vert \mathbf{p}_{i}-\mathbf{p}_{j}\right\Vert ^{2}=R$
will make the barrier function term unbounded. The separation distance
for robots indoor is often the sum of the two robots'{} physical
radius, namely $\left\Vert \mathbf{p}_{i}-\mathbf{p}_{j}\right\Vert ^{2}<R$
will not happen in practice. But, the separation distance is set largely
for UAVs compared with their sizes. Due to some uncertainties such
as communication delay, $\left\Vert \mathbf{p}_{i}-\mathbf{p}_{j}\right\Vert ^{2}<R$
will happen in the air. As a consequence, the control corresponding
to the barrier function terms mentioned above will make $\left\Vert \mathbf{p}_{i}-\mathbf{p}_{j}\right\Vert ^{2}\rightarrow0$
if $1\mathord{\left/\left(\left\Vert {\bf p}_{i}-{\bf p}_{j}\right\Vert ^{2}-R\right)\right.}$
is used (the two UAVs are pushed together by the design controller)
or appear numerical computation error if $\ln\left(\left\Vert \mathbf{p}_{i}-\mathbf{p}_{j}\right\Vert -R\right)$
is used. 
\end{itemize}
Motivated by these problems, practical distributed control is proposed
here to solve the virtual tube passing problem. Such a problem can
be classified into a \emph{basic virtual tube passing problem }and\emph{
}a\emph{ general virtual tube passing problem. }The former only considers
that all\textbf{ }UAVs are within the virtual tube at the beginning,
while the latter allows UAVs at everywhere in the beginning. For the
basic virtual tube passing problem,\emph{ }in light of artificial
potential field methods, one Lyapunov function and two barrier functions
are designed elaborately for approaching the finishing line, avoiding
conflict with other UAVs, and keeping within the virtual tube, respectively.
The distributed controller design is based on the combination of the
three Lyapunov-like functions. A formal proof is given to show that
all UAVs can pass the virtual tube without trapping, avoid conflict
and keep within the virtual tube. What is more, by the proposed control,
a UAV can keep away from another UAV or return back to the virtual
tube as soon as possible, once it enters into the safety area of another
UAV or has a conflict with the virtual tube during it is passing the
virtual tube. For the general virtual tube passing problem,\emph{
}several virtual tube type areas are defined to cover the whole airspace.
As a consequence, the general virtual tube passing problem is decomposed
into several basic\textbf{ }virtual tube passing problems. As a result,
for UAVs in different areas, they have different controllers according
to the design of the basic virtual tube passing problem. By switching
these controllers, the general virtual tube passing problem can be
solved.

The practicability of the proposed distributed control lies on the
following six features: 
\begin{itemize}
\item \emph{No ID required}. Unlike the formation control, neighboring UAVs'
IDs of a UAV are not required by the proposed distributed control.
Some active detection devices such as cameras or radars may only detect
neighboring UAVs' position and velocity but no IDs, because these
UAVs may look similarly. This implies that the proposed distributed
control can work autonomously without communication. 
\item \emph{Practical model used}. A double integral model with the given
velocity command as input is proposed for UAVs. This model is simple
and easy to obtain. What is more, distributed control is developed
for various tasks based on commercial semi-autonomous autopilots. 
\item \emph{Control saturation}. The maximum velocity command in the proposed
distributed controller is confined according to the requirement of
semi-autonomous autopilots. 
\item \emph{Conflict-free}. A formal proof about conflict avoidance and
keeping within the virtual tube is given. Even if a UAV enters into
the safety area of another UAV or has a conflict with the virtual
tube, it can keep away from the UAV or can return back to the virtual
tube as soon as possible. 
\item \emph{Convergence}. A formal proof is given to show that all UAVs
pass the finishing line without getting trapped. 
\item \emph{Low time complexity}. The proposed control protocol is simple
and can be computed at high speed, which is more suitable for dense
air traffic than optimization-based algorithms. The calculation time
of finding feasible solutions for different strategies will be compared
in simulation. 
\end{itemize}
The paper is organised as follows. In Section II, a UAV control model
is proposed, which contains the filtered position model and three
types of areas, to formulate the virtual tube passing problem composed
of a basic one and a general one. Some functions for different purposes
are introduced in Section III for controller design. In Section IV,
a controller is designed based on Lyapunov-like functions to solve
the basic virtual tube passing problem, where the stability analysis
is made. In Section V, a controller is designed based on Lyapunov-like
functions to solve the general virtual tube passing problem, by decomposing
this problem into several basic virtual tube passing problems. The
effectiveness of the proposed method is demonstrated by simulation
and flight experiments in Section VI. The conclusions are given in
Section VII. Some details of mathematical proof process are given
in Section VIII as appendix.

\section{Problem Formulation}

In this section, a UAV control model is introduced first, including
three types of areas, namely safety area, avoidance area, and detection
area, used for control. Then, the virtual tube passing\emph{ }problem
is formulated into a basic one and a general one depending on the
initial places of UAVs.

\subsection{UAV Control Model}

\subsubsection{Position Model}

There are ${M}$ VTOL UAVs in local airspace at the same altitude
satisfying the following model 
\begin{align}
\mathbf{\dot{p}}_{i} & =\mathbf{v}_{i}\nonumber \\
\mathbf{\dot{v}}_{i} & =-l_{i}\left(\mathbf{v}_{i}-\mathbf{v}_{\text{c},i}\right)\label{positionmodel_ab_con_i}
\end{align}
where $l_{i}>0,$ $\mathbf{p}_{i}\in{{\mathbb{R}}^{2}}$ and $\mathbf{v}_{i}\in{{\mathbb{R}}^{2}}$
are the position and velocity of the $i$th VTOL UAV, $\mathbf{v}_{\text{c},i}\in{{\mathbb{R}}^{2}}$
is the velocity command of the $i$th UAV, $i=1,2,\cdots,M.$ The
control gain $l_{i}$ depends on the $i$th UAV and the semi-autonomous
autopilot used, which can be obtained through flight experiments.
From the model (\ref{positionmodel_ab_con_i}), $\lim_{t\rightarrow\infty}\left\Vert \mathbf{v}_{i}\left(t\right)-\mathbf{v}_{\text{c},i}\right\Vert =0$
if $\mathbf{v}_{\text{c},i}$ is constant. Here, the velocity command
$\mathbf{v}_{\text{c},i}$ for the $i$th VTOL UAV, is subject to
a saturation defined as where ${v_{\text{m},i}}>0$ is the maximum
speed of the $i$th VTOL UAVs, $i=1,2,\cdots,M$, $\mathbf{v}\triangleq\lbrack{{v}_{1}}$
${{v}_{2}}]{^{\text{T}}}\in{{\mathbb{R}}^{2}}$. The saturation function
sa${\text{t}}\left(\mathbf{v},{v_{\text{m},i}}\right)$ and the vector
$\mathbf{v}$ are parallel all the time so it can keep the flying
direction the same if $\left\Vert \mathbf{v}\right\Vert >{v_{\text{m},i}}$
\cite{Quan(2017)}. The saturation function can be rewritten as 
\begin{equation}
\text{sat}\left(\mathbf{v},{v_{\text{m},i}}\right)={{\kappa}_{{v_{\text{m},i}}}}\left(\mathbf{v}\right)\mathbf{v}\label{sat0}
\end{equation}
where 
\[
{{\kappa}_{{v_{\text{m},i}}}}\left(\mathbf{v}\right)\triangleq\left\{ \begin{array}{c}
1,\\
\frac{{v_{\text{m},i}}}{\left\Vert \mathbf{v}\right\Vert },
\end{array}\begin{array}{c}
\left\Vert \mathbf{v}\right\Vert \leq{v_{\text{m},i}}\\
\left\Vert \mathbf{v}\right\Vert >{v_{\text{m},i}}
\end{array}\right..
\]
It is obvious that $0<{{\kappa}_{{v_{\text{m},i}}}}\left(\mathbf{v}\right)\leq1$.
Sometimes, ${{\kappa}_{{v_{\text{m},i}}}}\left(\mathbf{v}\right)$
will be written as ${{\kappa}_{{v_{\text{m},i}}}}$ for short. According
to this, if and only if $\mathbf{v=0},$ then 
\begin{equation}
\mathbf{v}^{\text{T}}\text{sa}{\text{t}}\left(\mathbf{v},{v_{\text{m},i}}\right)=0.\label{saturation1}
\end{equation}

\textbf{Remark 1. }It is well-known that a typical multicopter is
a physical system with underactuated dynamics \cite[pp.126-130]{Quan(2017)}.
But, many organizations or companies have designed some open source
semi-autonomous autopilots or offered semi-autonomous autopilots with
software development kits. The semi-autonomous autopilots can be used
for velocity control of VTOL UAVs. For example, A3 autopilots released
by DJI allow the range of the horizontal velocity command from $-10$m/s$\sim10$m/s
\cite{A3}. With such an autopilot, the velocity of a VTOL UAV can
track a given velocity command in a reasonable time. Not only can
this avoid the trouble of modifying the low-level source code of autopilots,
but also it can utilize commercial autopilots to complete various
tasks. So, the dynamics (\ref{positionmodel_ab_con_i}) is practical
especially for higher-level control.

\subsubsection{Filtered Position Model}

\begin{figure}[h]
\begin{centering}
\includegraphics[scale=0.9]{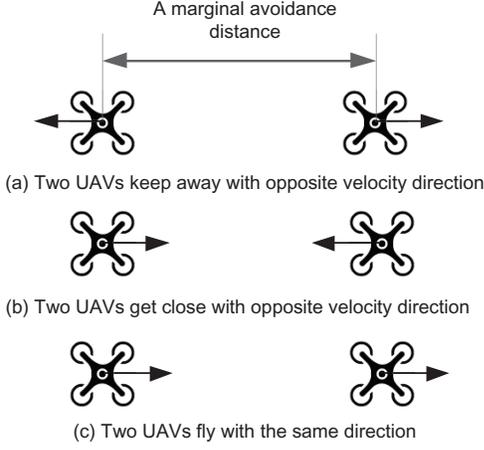} 
\par\end{centering}
\caption{Intuitive interpretation for filtered position}
\label{Intuitive} 
\end{figure}

In this section, the motion of each VTOL UAV is transformed into a
single integrator form to simplify the controller design and analysis.
As shown in Figure \ref{Intuitive}, although the position distances
of the three cases are the same, namely a marginal avoidance distance,
the case in Figure \ref{Intuitive}(b) needs to carry out avoidance
urgently by considering the velocity. However, the case in Figure
\ref{Intuitive}(a) in fact does not need to be considered to perform
collision avoidance. With such an intuition, a filtered position is
defined as follows: 
\begin{equation}
\boldsymbol{\xi}_{i}\triangleq{\mathbf{p}}_{i}+\frac{1}{{l}_{i}}\mathbf{v}_{i}.\label{FilteredPosition}
\end{equation}
{Then} 
\begin{align}
\boldsymbol{\dot{\xi}}_{i} & =\mathbf{\dot{p}}_{i}+\frac{1}{{l}_{i}}\mathbf{\dot{v}}_{i}\nonumber \\
 & =\mathbf{v}_{i}-\frac{1}{{l}_{i}}{l}_{i}\left(\mathbf{v}_{i}-\mathbf{v}_{\text{c},i}\right)\nonumber \\
 & =\mathbf{v}_{\text{c},i}\label{filteredposdyn}
\end{align}
where $i=1,2,\cdots,M${. Let} 
\begin{equation}
r_{\text{v}}=\max_{i}\frac{v_{\text{m},i}}{l_{i}}.\label{rv}
\end{equation}

In the following, a relationship between the position error and the
filtered position error is shown.

\textbf{Proposition 1}. Given any $r>0,$ for the $i$th and $j$th
VTOL UAVs, if $\left\Vert \mathbf{v}_{i}\left(0\right)\right\Vert \leq{v_{\text{m},i}}$
and the filtered position error satisfies $\left\Vert \boldsymbol{\xi}_{i}\left(t\right)-\boldsymbol{\xi}_{j}\left(t\right)\right\Vert \geq r+2r_{\text{v}},$
then 
\[
\left\Vert \mathbf{p}_{i}\left(t\right)-{{\mathbf{p}}_{j}}\left(t\right)\right\Vert \geq r
\]
$t\geq0,$ where $i,j=1,2,\cdots,M,$ $i\neq j,\ r>0$.

\emph{Proof}. See \emph{Appendix}. $\square$

\subsection{Three Types of Areas around a UAV}

In light of \cite{Connectedness}, three types of areas used for control,
namely safety area, avoidance area, and detection area, are defined.
Unlike \cite{Connectedness}, these areas are suit for UAVs and the
velocity is further introduced.

\begin{figure}[h]
\begin{centering}
\includegraphics{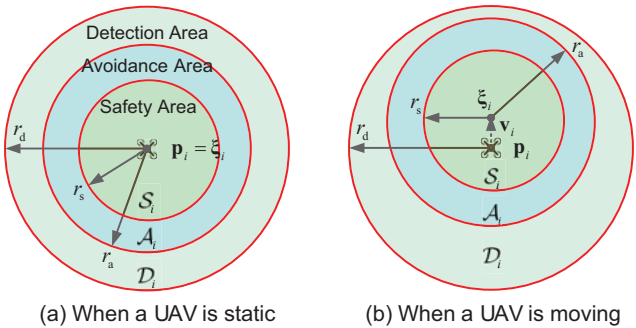} 
\par\end{centering}
\caption{Safety area, avoidance area and detection area of a UAV.}
\label{Threeaera} 
\end{figure}

\subsubsection{Safety Area}

In order to avoid a conflict, as shown in Figure \ref{Threeaera},
the \emph{safety} \emph{radius} $r_{\text{s}}$ of a UAV is defined
as 
\begin{equation}
\mathcal{S}_{i}=\left\{ \mathbf{x}\in{{\mathbb{R}}^{2}}\left\vert \left\Vert \mathbf{x}-\boldsymbol{\xi}_{i}\right\Vert \leq r_{\text{s}}\right.\right\} \label{Safetyaera}
\end{equation}
where $r_{\text{s}}>0$ is the safety radius, $i=1,2,\cdots,M.$ It
should be noted that we consider the velocity of the $i$th UAV in
the definition of $\mathcal{S}_{i}$. For all UAVs, no \emph{confliction}
with each other implies 
\[
\mathcal{S}_{i}\cap\mathcal{S}_{j}=\varnothing
\]
namely 
\begin{equation}
\left\Vert \boldsymbol{\xi}_{j}-\boldsymbol{\xi}_{i}\right\Vert >2r_{\text{s}}.\label{sdis}
\end{equation}
\textit{Proposition 1} implies that two VTOL UAVs will be separated
largely enough if (\ref{sdis}) is satisfied with a safety radius\emph{
}$r_{\text{s}}$ large enough.

\subsubsection{Avoidance Area}

Besides the safety area, there exists an \emph{avoidance area} used
for starting avoidance control. If another UAV is out of the avoidance
area of the $i$th UAV, then the object will not need to be avoided.
For the $i$th UAV, the \emph{avoidance area }for other UAVs is\emph{
}defined as 
\begin{equation}
\mathcal{A}_{i}=\left\{ \mathbf{x}\in{{\mathbb{R}}^{2}}\left\vert \left\Vert \mathbf{x}-\boldsymbol{\xi}_{i}\right\Vert \leq r_{\text{a}}\right.\right\} \label{Avoidancearea}
\end{equation}
where $r_{\text{a}}>0$ is the \emph{avoidance radius}, $i=1,2,\cdots,M.$
It should be noted that we consider the velocity of the $i$th UAV
in the definition of $\mathcal{A}_{i}$. If 
\[
\mathcal{A}_{i}\cap\mathcal{S}_{j}\neq\varnothing,
\]
namely 
\[
\left\Vert \boldsymbol{\xi}_{i}-\boldsymbol{\xi}_{j}\right\Vert \leq r_{\text{a}}+r_{\text{s}}
\]
then the $j$th UAV should be avoided by the $i$th UAV. Since 
\[
\mathcal{A}_{i}\cap\mathcal{S}_{j}\neq\varnothing\Leftrightarrow\mathcal{A}_{j}\cap\mathcal{S}_{i}\neq\varnothing
\]
according to the definition of $\mathcal{A}_{i},$ the $i$th UAV
should be avoided by the $j$th UAV at the same time. When the $j$th
UAV just enters into the avoidance area of the $i$th UAV, it is required
that they have not conflicted at the beginning. Therefore,\textbf{
}we require 
\[
r_{\text{a}}>r_{\text{s}}.
\]

\subsubsection{Detection Area}

By cameras, radars, or Vehicle to Vehicle (V2V) communication, the
UAVs can receive the positions and velocities of their neighboring
UAVs. The \emph{detection area }only\emph{ }depends on the detection
range of the used devices, which is only related to its position.
For the $i$th UAV, this area is defined as 
\begin{equation}
\mathcal{D}_{i}=\left\{ \mathbf{x}\in\mathbb{R}^{2}\left\vert \left\Vert \mathbf{x}-\mathbf{p}_{i}\right\Vert \leq r_{\text{d}}\right.\right\} \label{DetectionArea}
\end{equation}
where $r_{\text{d}}>0$ is the \emph{detection radius},$\ i=1,2,\cdots,M$.
When another UAV is within this area, it can be detected.

\textbf{Proposition 2}. Suppose $r_{\text{d}}>r_{\text{s}}+r_{\text{a}}+2r_{\text{v}}{,}$
$i=1,2,\cdots,M{.}$ Then for any $i\neq j,$ if $\mathcal{A}_{i}\cap\mathcal{S}_{j}\neq\varnothing,$
then $\mathbf{p}_{j}\in\mathcal{D}_{i},$ $i,j=1,2,\cdots,M.$

\textit{Proof}. It is similar to proof of \textit{Proposition 1. }$\square$

To simplify the following problems, we have the following assumption
for all VTOL UAVs.

\textbf{Assumption 1}. The radius of the detection area satisfies
$r_{\text{d}}>r_{\text{s}}+r_{\text{a}}+2r_{\text{v}}{.}$

According to \textit{Assumption 1},\textbf{ }for the $i$th UAV, any
other UAV entering into its avoidance area can be detected by the
$i$th UAV and will not conflict with the $i$th UAV initially, $i=1,2,\cdots,M{.}$

\subsection{virtual tube Passing\emph{ }Problem Formulation}

\begin{figure}[h]
\begin{centering}
\includegraphics{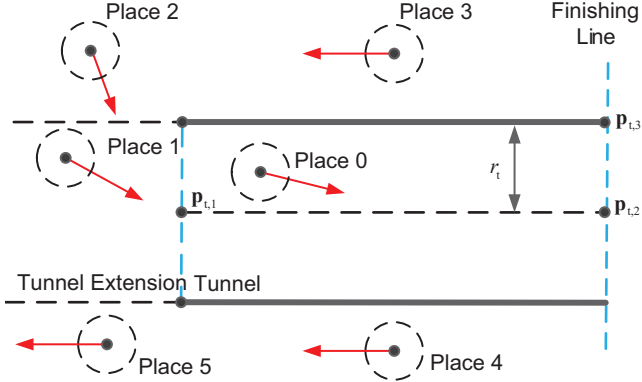} 
\par\end{centering}
\caption{Airspace and virtual tube.}
\label{airwaytunnel} 
\end{figure}

In a horizontal plane, as shown in Figure \ref{airwaytunnel}, a virtual
tube (analogous to an \emph{airway} or a \emph{highway} on the ground)
here is a horizontal long band with the width $2r_{\text{t}}$ and
centerline starting from $\mathbf{p}_{\text{t,1}}\in\mathbb{R}^{2}$
to $\mathbf{p}_{\text{t,2}}\in\mathbb{R}^{2},$ where $r_{\text{t}}>L{{r}_{\text{a}}},$
where $L\in{\mathbb{
\mathbb{Z}
}}_{+}$ is the lane number in the virtual tube allowed for UAVs. 

\begin{figure}[h]
\begin{centering}
\includegraphics{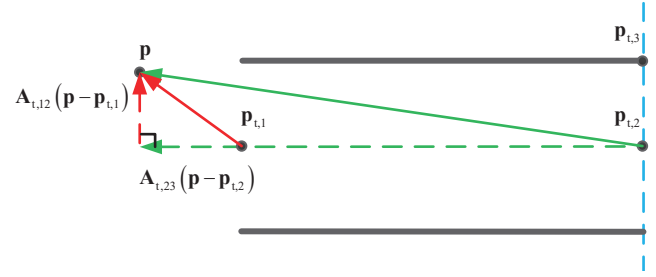} 
\par\end{centering}
\caption{Diagram of the projective operator.}
\label{Projection} 
\end{figure}

Define 
\begin{align}
\mathbf{A}_{\text{t,12}}\left(\mathbf{p}_{\text{t,1}},\mathbf{p}_{\text{t,2}}\right) & \triangleq\mathbf{I}_{2}-\frac{\left(\mathbf{p}_{\text{t,1}}-\mathbf{p}_{\text{t,2}}\right)\left(\mathbf{p}_{\text{t,1}}-\mathbf{p}_{\text{t,2}}\right){}^{\text{T}}}{\left\Vert \mathbf{p}_{\text{t,1}}-\mathbf{p}_{\text{t,2}}\right\Vert ^{2}}\nonumber \\
\mathbf{A}_{\text{t,23}}\left(\mathbf{p}_{\text{t,2}},\mathbf{p}_{\text{t,3}}\right) & \triangleq\mathbf{I}{}_{2}-\frac{\left(\mathbf{p}_{\text{t,2}}-\mathbf{p}_{\text{t,3}}\right)\left(\mathbf{p}_{\text{t,2}}-\mathbf{p}_{\text{t,3}}\right){}^{\text{T}}}{\left\Vert \mathbf{p}_{\text{t,2}}-\mathbf{p}_{\text{t,3}}\right\Vert ^{2}}.\label{define}
\end{align}
Here, matrix $\mathbf{A}_{\text{t,12}}=\mathbf{A}_{\text{t,12}}^{\text{T}},$
$\mathbf{A}_{\text{t,23}}=\mathbf{A}_{\text{t,23}}^{\text{T}}\in{\mathbb{R}}^{2\times2}$
are positive semi-definite matrices. According to the projection operator
\cite[p. 480]{Projection}, the value $\left\Vert \mathbf{A}_{\text{t,12}}\left(\mathbf{p}-{{\mathbf{p}}_{\text{t,1}}}\right)\right\Vert $
is the distance from $\mathbf{p\in
\mathbb{R}
}^{2}$ to the straight line $\overline{{{\mathbf{p}}_{\text{t,1}}{\mathbf{p}}_{\text{t,2}}}}$
as shown in Figure \ref{Projection}. Particularly, the equation $\left\Vert \mathbf{A}_{\text{t,12}}\left(\mathbf{p}-{{\mathbf{p}}_{\text{t,1}}}\right)\right\Vert =0$
implies that $\mathbf{p}$ is on the straight-line $\overline{{{\mathbf{p}}_{\text{t,1}}{\mathbf{p}}_{\text{t,2}}}}$.
Similarly, the value $\left\Vert \mathbf{A}_{\text{t,23}}\left(\mathbf{p}-{{\mathbf{p}}_{\text{t,2}}}\right)\right\Vert $
is the distance from $\mathbf{p}$ to the finishing line $\overline{{{\mathbf{p}}_{\text{t,2}}{\mathbf{p}}_{\text{t,3}}}}$.
Define position errors as 
\begin{align*}
\mathbf{\tilde{p}}{_{\text{l,}i}} & \triangleq\mathbf{A}_{\text{t,23}}\left(\mathbf{p}_{i}-{{\mathbf{p}}_{\text{t,2}}}\right)\\
\mathbf{\tilde{p}}{_{\text{m,}ij}} & \triangleq\mathbf{p}_{i}-{{\mathbf{p}}_{j}}\\
\mathbf{\tilde{p}}{_{\text{t,}i}} & \triangleq\mathbf{A}_{\text{t,12}}\left(\mathbf{p}_{i}-{{\mathbf{p}}_{\text{t,2}}}\right)
\end{align*}
and the filtered position errors as 
\begin{align*}
\boldsymbol{\tilde{\xi}}{}_{\text{l,}i} & \triangleq\mathbf{A}_{\text{t,23}}\left(\boldsymbol{\xi}_{i}-\mathbf{p}_{\text{t,2}}\right)\\
\boldsymbol{\tilde{\xi}}{}_{\text{m,}ij} & \triangleq\boldsymbol{\xi}_{i}-\boldsymbol{\xi}{}_{j}\\
\boldsymbol{\tilde{\xi}}{}_{\text{t,}i} & \triangleq\mathbf{A}_{\text{t,12}}\left(\boldsymbol{\xi}_{i}-\mathbf{p}_{\text{t,2}}\right)
\end{align*}
where $i,j=1,2,\cdots,M.$ With the definitions above, according to
(\ref{positionmodel_ab_con_i}), the derivatives of the filtered errors
are 
\begin{align}
\boldsymbol{\dot{\tilde{\xi}}}{_{\text{l,}i}} & =\mathbf{A}_{\text{t,23}}\mathbf{v}_{\text{c},i}\label{lmodel}\\
\boldsymbol{\dot{\tilde{\xi}}}{_{\text{m,}ij}} & =\mathbf{v}_{\text{c},i}-\mathbf{v}_{\text{c},j}\label{mmodel}\\
\boldsymbol{\dot{\tilde{\xi}}}{}_{\text{t,}i} & =\mathbf{A}_{\text{t,12}}\mathbf{v}_{\text{c},i}\label{tmodel}
\end{align}
where $i,j=1,2,\cdots,M.$

With the description above, the following assumptions are proposed.

\textbf{Assumption 2}. As shown in Figure \ref{airwaytunnel}, the
initial condition $\mathbf{p}_{i}\left(0\right),\boldsymbol{\xi}_{i}\left(0\right),$
$i=1,2,\cdots,M$ are all within the virtual tube or its extension,
namely 
\begin{align*}
\left(\frac{{{\mathbf{p}}_{\text{t,2}}}-{{\mathbf{p}}_{\text{t,1}}}}{\left\Vert {{\mathbf{p}}_{\text{t,2}}}-{{\mathbf{p}}_{\text{t,1}}}\right\Vert }\right)^{\text{T}}\left(\mathbf{p}_{i}\left(0\right)-{{\mathbf{p}}_{\text{t,2}}}\right) & <0\\
\left(\frac{{{\mathbf{p}}_{\text{t,2}}}-{{\mathbf{p}}_{\text{t,1}}}}{\left\Vert {{\mathbf{p}}_{\text{t,2}}}-{{\mathbf{p}}_{\text{t,1}}}\right\Vert }\right)^{\text{T}}\left(\boldsymbol{\xi}_{i}\left(0\right)-{{\mathbf{p}}_{\text{t,2}}}\right) & <0
\end{align*}
where $\overline{{{\mathbf{p}}_{\text{t,2}}{\mathbf{p}}_{\text{t,3}}}}$
is perpendicular to $\overline{{{\mathbf{p}}_{\text{t,1}}{\mathbf{p}}_{\text{t,2}}}}$
with $\left\Vert {{\mathbf{p}}_{\text{t,2}}}-{{\mathbf{p}}_{\text{t,3}}}\right\Vert =r_{\text{t}}.$

\textbf{Assumption 2}$^{\prime}$. As shown in Figure \ref{airwaytunnel},
the initial condition $\boldsymbol{\xi}_{i}\left(0\right)$ are not
all within the virtual tube or its extension, but locate the left
of the finishing line $\overline{{{\mathbf{p}}_{\text{t,2}}{\mathbf{p}}_{\text{t,3}}}}$.

\textbf{Assumption 3}. The UAVs' initial conditions satisfy 
\[
\left\Vert \boldsymbol{\xi}_{i}\left(0\right)-\boldsymbol{\xi}_{j}\left(0\right)\right\Vert >2r_{\text{s}},i\neq j
\]
and $\left\Vert \mathbf{v}_{i}\left(0\right)\right\Vert \leq{v_{\text{m}},}$
where $i,j=1,2,\cdots,M.$

\textbf{Assumption 4}. Once a UAV arrives near the finishing line
$\overline{{{\mathbf{p}}_{\text{t,2}}{\mathbf{p}}_{\text{t,3}}}}$,
then it will quit the virtual tube not to affect the UAVs behind.
Mathematically, given ${\epsilon}_{\text{0}}\in
\mathbb{R}
_{+},$ a UAV arrives near the finishing line $\overline{{{\mathbf{p}}_{\text{t,2}}{\mathbf{p}}_{\text{t,3}}}}$
if 
\begin{equation}
{\left({{\mathbf{p}}_{\text{t,2}}}-{{\mathbf{p}}_{\text{t,1}}}\right){^{\text{T}}}}\mathbf{A}_{\text{t,23}}\left(\mathbf{p}_{i}-{{\mathbf{p}}_{\text{t,2}}}\right)\geq-{\epsilon}_{\text{0}}.\label{arrivialairway}
\end{equation}

\begin{figure}[h]
\begin{centering}
\includegraphics{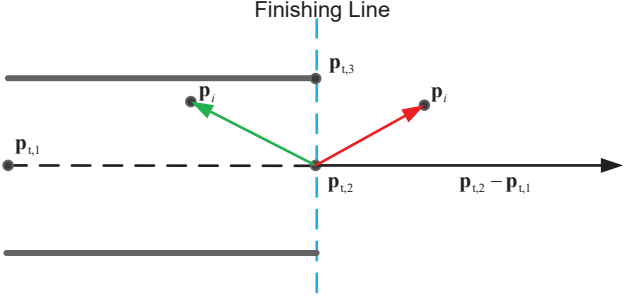} 
\par\end{centering}
\caption{Position relative to the finishing line.}
\label{Remark} 
\end{figure}

\textbf{Neighboring Set}. Let the set $\mathcal{N}_{\text{m},i}$
be the collection of all mark numbers of other VTOL UAVs whose safety
areas enter into the avoidance\emph{ }area of the $i$th UAV, namely
\[
\mathcal{N}_{\text{m},i}=\left\{ \left.j\right\vert \mathcal{S}_{j}\cap\mathcal{A}_{i}\neq\varnothing,j=1,\cdots,M,i\neq j\right\} .
\]
For example, if the safety areas of the $1$th, $2$th VTOL UAVs enter
into in the avoidance\emph{ }area of the $3$th UAV, then $\mathcal{N}_{\text{m},3}=\left\{ 1,2\right\} $.
Based on \textit{Assumptions} and \textit{definition} above, two types
of \emph{virtual tube passing problems} are stated in the following. 
\begin{itemize}
\item \textbf{Basic virtual tube passing problem}.\emph{ }Under \textit{Assumptions
1-4}, design the velocity input $\mathbf{v}_{\text{c},i}$ for the
$i$th UAV with local information from $\mathcal{N}_{\text{m},i}$
to guide it to fly to pass the virtual tube until it arrives near
the finishing line $\overline{{{\mathbf{p}}_{\text{t,2}}{\mathbf{p}}_{\text{t,3}}}}$,
meanwhile avoiding colliding other UAVs ($\left\Vert \boldsymbol{\tilde{\xi}}{_{\text{m,}ij}}\right\Vert >2r_{\text{s}}$)
and keeping within the virtual tube ($\left\Vert \boldsymbol{\tilde{\xi}}{_{\text{t,}i}}\right\Vert <r_{\text{t}}-r_{\text{s}}$)
while passing it, $i=1,2,\cdots,M$. 
\item \textbf{General virtual tube passing problem}.\emph{ }Under \textit{Assumptions
1,2}$^{\prime}$\textit{,3,4}, design the velocity input $\mathbf{v}_{\text{c},i}$
for the $i$th UAV with local information from $\mathcal{N}_{\text{m},i}$
to guide it to fly to pass the virtual tube until it arrives near
the finishing line $\overline{{{\mathbf{p}}_{\text{t,2}}{\mathbf{p}}_{\text{t,3}}}}$,
meanwhile avoiding conflict with other UAVs ($\left\Vert \boldsymbol{\tilde{\xi}}{_{\text{m,}ij}}\right\Vert >2r_{\text{s}}$)
and keeping within the virtual tube when passing it ($\left\Vert \boldsymbol{\tilde{\xi}}{_{\text{t,}i}}\right\Vert <r_{\text{t}}-r_{\text{s}}$),
$i=1,2,\cdots,M$. 
\end{itemize}
\textbf{Remark 1}. As shown in Figure \ref{Remark}, if $\left({{\mathbf{p}}_{\text{t,2}}}-{{\mathbf{p}}_{\text{t,1}}}\right)^{\text{T}}\left(\mathbf{p}_{i}-{{\mathbf{p}}_{\text{t,2}}}\right)>{0,}$
then $\mathbf{p}{_{i}}$ locates the right side of the finishing line
$\overline{{{\mathbf{p}}_{\text{t,2}}{\mathbf{p}}_{\text{t,3}}}}$;
if $\left({{\mathbf{p}}_{\text{t,2}}}-{{\mathbf{p}}_{\text{t,1}}}\right)^{\text{T}}\left(\mathbf{p}_{i}-{{\mathbf{p}}_{\text{t,2}}}\right){<0,}$
then $\mathbf{p}{_{i}}$ locates the left side of the finishing line
$\overline{{{\mathbf{p}}_{\text{t,2}}{\mathbf{p}}_{\text{t,3}}}}$.

\textbf{Remark 2}. For \textit{Assumption 2},\textbf{ }all\textbf{
}UAVs are within the virtual tube (like \textit{Place 0 }in Figure
\ref{airwaytunnel}) or its extension (like \textit{Place 1 }in Figure
\ref{airwaytunnel}). For \textit{Assumption 2}$^{\prime}$,\textbf{
}UAVs are not all within the virtual tube or its extension. This implies
that UAVs may locate everywhere. For example, UAVs may locate the
places, like \textit{Place 0}, ..., \textit{Place 5} shown in Figure
\ref{airwaytunnel}.

\section{Preliminaries}

\subsection{Line Integral Lyapunov Function}

In the following, we will design a new type of Lyapunov functions,
called \emph{Line Integral Lyapunov Function. }This type of Lyapunov
functions is inspired by its scalar form \cite[p.74]{Slotine(1991)}.
If $xf\left(x\right)>0\ $for $x\neq0,$ then $V_{\text{li}}^{\prime}\left(y\right)=\int_{0}^{y}f\left(x\right)$d$x>0$
when $y\neq0.$ The derivative is $\dot{V}_{\text{li}}^{\prime}=f\left(y\right)\dot{y}.$
A line integral Lyapunov function for vectors is defined as 
\begin{equation}
V_{\text{li}}\left(\mathbf{y}\right)=\int_{C_{\mathbf{y}}}\text{sa}{\text{t}}\left(\mathbf{x},a\right)^{\text{T}}\text{d}\mathbf{x}\label{Vli0}
\end{equation}
where $a>0,$ $\mathbf{x}\in\mathbf{
\mathbb{R}
}^{n},$ $C_{\mathbf{y}}$ is a line from $\mathbf{0}$ to $\mathbf{y}\in
\mathbb{R}
^{n}\mathbf{.}$ In the following lemma, we will show its properties.

\textbf{Lemma 1}. Suppose that the line integral Lyapunov function
$V_{\text{li}}$ is defined as (\ref{Vli0}). Then (i) $V_{\text{li}}\left(\mathbf{y}\right)>0$
if $\left\Vert \mathbf{y}\right\Vert \neq0$; (ii) if $\left\Vert \mathbf{y}\right\Vert \rightarrow\infty,$
then $V_{\text{li}}\left(\mathbf{y}\right)\rightarrow\infty;$ (iii)
if $V_{\text{li}}\left(\mathbf{y}\right)$ is bounded, then $\left\Vert \mathbf{y}\right\Vert $
is bounded.

\textit{Proof}. Since 
\[
\text{sa{t}}\left(\mathbf{x},{a}\right)={{\kappa}_{{a}}}\left(\mathbf{x}\right)\mathbf{x}
\]
the function (\ref{Vli0}) can be written as 
\begin{equation}
V_{\text{li}}\left(\mathbf{y}\right)=\int_{C_{\mathbf{y}}}{{\kappa}_{{a}}}\left(\mathbf{x}\right)\mathbf{x}^{\text{T}}\text{d}\mathbf{x}\label{Vli10}
\end{equation}
where 
\[
{{\kappa}_{{a}}}\left(\mathbf{x}\right)\triangleq\left\{ \begin{array}{c}
1,\\
\frac{{a}}{\left\Vert \mathbf{x}\right\Vert },
\end{array}\begin{array}{c}
\left\Vert \mathbf{x}\right\Vert \leq{a}\\
\left\Vert \mathbf{x}\right\Vert >{a}
\end{array}\right..
\]
Let $z=\left\Vert \mathbf{x}\right\Vert .$ Then the function (\ref{Vli10})
becomes 
\begin{align*}
V_{\text{li}}\left(\mathbf{y}\right) & =\int_{C_{\mathbf{y}}}\frac{{{\kappa}_{{a}}}\left(\mathbf{x}\right)}{2}\text{d}z^{2}\\
 & =\int_{0}^{\left\Vert \mathbf{y}\right\Vert }{{\kappa}_{{a}}}\left(\mathbf{x}\right)z\text{d}z.
\end{align*}

\begin{itemize}
\item If $\left\Vert \mathbf{y}\right\Vert \leq{a,}$ then ${{\kappa}_{{a}}}\left(\mathbf{x}\right)=1.$
Consequently, 
\begin{equation}
V_{\text{li}}\left(\mathbf{y}\right)=\frac{1}{2}\left\Vert \mathbf{y}\right\Vert ^{2}.\label{Vli2}
\end{equation}
\item If $\left\Vert \mathbf{y}\right\Vert >{a,}$ then 
\[
\int_{0}^{\left\Vert \mathbf{y}\right\Vert }{{\kappa}_{{a}}}\left(\mathbf{x}\right)z\text{d}z=\int_{0}^{a}z\text{d}z+\int_{a}^{\left\Vert \mathbf{y}\right\Vert }\frac{{a}}{\left\Vert \mathbf{x}\right\Vert }z\text{d}z.
\]
Since $z=\left\Vert \mathbf{x}\right\Vert ,$ we have 
\begin{equation}
V_{\text{li}}\left(\mathbf{y}\right)\geq\frac{1}{2}a^{2}+{a}\left(\left\Vert \mathbf{y}\right\Vert -a\right).\label{Vli3}
\end{equation}
\end{itemize}
Therefore, from the form of (\ref{Vli2}) and (\ref{Vli3}), we have
(i) $V_{\text{li}}\left(\mathbf{y}\right)>0$ if $\left\Vert \mathbf{y}\right\Vert \neq0$.
(ii) if $\left\Vert \mathbf{y}\right\Vert \rightarrow\infty,$ then
$V_{\text{li}}\left(\mathbf{y}\right)\rightarrow\infty;$ (iii) if
$V_{\text{li}}\left(\mathbf{y}\right)$ is bounded, then $\left\Vert \mathbf{y}\right\Vert $
is bounded. $\square$

\subsection{Two Smooth Functions}

\begin{figure}[h]
\begin{centering}
\includegraphics{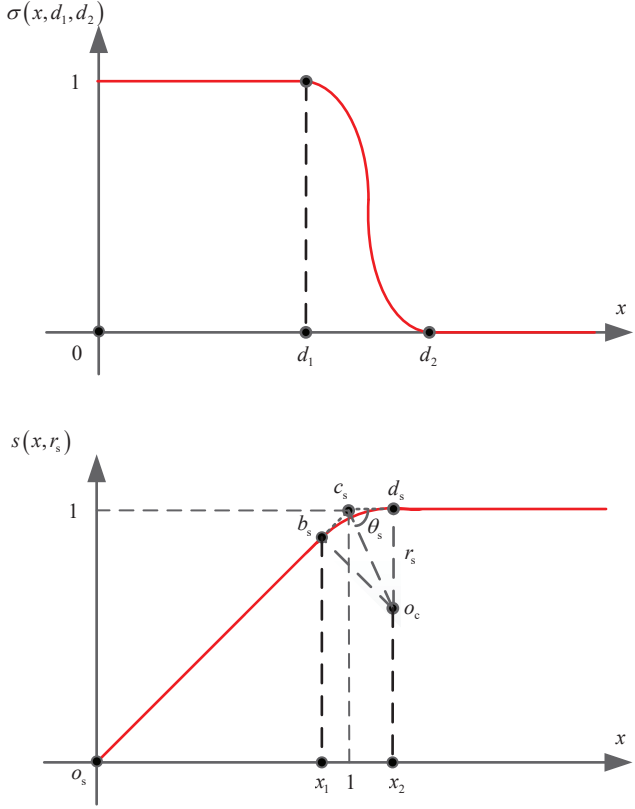} 
\par\end{centering}
\caption{Two smooth functions. For a smooth saturation function, $\theta_{\text{s}}=67.5^{\circ}.$}
\label{saturationa} 
\end{figure}

Two smooth functions are defined for the following Lyapunov-like function
design. As shown in Figure \ref{saturationa} (upper plot), define
a second-order differentiable `bump' function as \cite{Panagou(2016)}
\begin{equation}
\sigma\left(x,d_{1},d_{2}\right)=\left\{ \begin{array}{c}
1\\
Ax^{3}+Bx^{2}+Cx+D\\
0
\end{array}\right.\begin{array}{c}
\text{if}\\
\text{if}\\
\text{if}
\end{array}\begin{array}{c}
x\leq d_{1}\\
d_{1}\leq x\leq d_{2}\\
d_{2}\leq x
\end{array}\label{zerofunction}
\end{equation}
with $A=-2\left/\left(d_{1}-d_{2}\right)^{3}\right.,$ $B=3\left(d_{1}+d_{2}\right)\left/\left(d_{1}-d_{2}\right)^{3}\right.,$
$C=-6d_{1}d_{2}\left/\left(d_{1}-d_{2}\right)^{3}\right.$ and $D=d_{2}^{2}\left(3d_{1}-d_{2}\right)\left/\left(d_{1}-d_{2}\right)^{3}\right.$.
The derivative of $\sigma\left(x,d_{1},d_{2}\right)$ with respect
to $x$ is 
\[
\frac{\partial\sigma\left(x,d_{1},d_{2}\right)}{\partial x}=\left\{ \begin{array}{c}
0\\
3Ax^{2}+2Bx+C\\
0
\end{array}\right.\begin{array}{c}
\text{if}\\
\text{if}\\
\text{if}
\end{array}\begin{array}{c}
x\leq d_{1}\\
d_{1}\leq x\leq d_{2}\\
d_{2}\leq x
\end{array}.
\]

Define another smooth function as shown in Figure \ref{saturationa}
(lower plot) to approximate a saturation function 
\[
\bar{s}\left(x\right)=\min\left(x,1\right),x\geq0
\]
that 
\begin{equation}
s\left(x,\epsilon_{\text{s}}\right)=\left\{ \begin{array}{c}
x\\
\left(1-\epsilon_{\text{s}}\right)+\sqrt{\epsilon_{\text{s}}^{2}-\left(x-x_{2}\right)^{2}}\\
1
\end{array}\right.\begin{array}{c}
0\leq x\leq x_{1}\\
x_{1}\leq x\leq x_{2}\\
x_{2}\leq x
\end{array}\label{sat}
\end{equation}
with $x_{2}=1+\frac{1}{\tan67.5^{\circ}}\epsilon_{\text{s}}$ and
$x_{1}=x_{2}-\sin45^{\circ}\epsilon_{\text{s}}.$ Since it is required
$x_{1}\geq0$, one has $\epsilon_{\text{s}}\leq\frac{\tan67.5^{\circ}}{\tan67.5^{\circ}\sin45^{\circ}-1}.$
For any $\epsilon_{\text{s}}\in\left[0,\frac{\tan67.5^{\circ}}{\tan67.5^{\circ}\sin45^{\circ}-1}\right],$
it is easy to see 
\begin{equation}
s\left(x,\epsilon_{\text{s}}\right)\leq\bar{s}\left(x\right)\label{satinequ}
\end{equation}
and 
\begin{equation}
\lim_{\epsilon_{\text{s}}\rightarrow0}\underset{x\geq0}{\sup}\left\vert \bar{s}\left(x\right)-s\left(x,\epsilon_{\text{s}}\right)\right\vert =0.\label{sata}
\end{equation}
The derivative of $s\left(x,\epsilon_{\text{s}}\right)$ with respect
to $x$ is 
\[
\frac{\partial s\left(x,\epsilon_{\text{s}}\right)}{\partial x}=\left\{ \begin{array}{c}
1\\
\frac{x_{2}-x}{\sqrt{\epsilon_{\text{s}}^{2}-\left(x-x_{2}\right)^{2}}}\\
0
\end{array}\right.\begin{array}{c}
0\leq x\leq x_{1}\\
x_{1}\leq x\leq x_{2}\\
x_{2}\leq x
\end{array}.
\]
For any $\epsilon_{\text{s}}>0,$ we have $\underset{x\geq0}{\sup}\left\vert \partial s\left(x,\epsilon_{\text{s}}\right)\left/\partial x\right.\right\vert \leq1.$

\section{Basic virtual tube Passing Problem}

In this section, three Lyapunov-like functions for approaching the
finishing line, avoiding conflict, and keeping within the virtual
tube are established. Based on them, a controller to solve the basic\textbf{
}virtual tube passing problem is derived and then a formal analysis
is made.

\subsection{Lyapunov-Like Function Design and Analysis}

For the basic\textbf{ }virtual tube passing problem, three subproblems
are required to solve, namely approaching the finishing line $\overline{{{\mathbf{p}}_{\text{t,2}}{\mathbf{p}}_{\text{t,3}}}}$,
avoiding conflict with other UAVs, and keeping within the virtual
tube. Correspondingly, three Lyapunov-like functions are proposed.

\subsubsection{Integral Lyapunov Function for Approaching Finishing Line}

Define a smooth curve $C_{\boldsymbol{\tilde{\xi}}{_{\text{l,}i}}}$
from $\mathbf{0}$ to $\boldsymbol{\tilde{\xi}}{_{\text{l,}i}}$.
Then, the line integral of sa${\text{t}}\left(\mathbf{x},{v_{\text{m},i}}\right)$
along $C_{\boldsymbol{\tilde{\xi}}{_{\text{l,}i}}}$ is 
\begin{equation}
V_{\text{l},i}=\int_{C_{\boldsymbol{\tilde{\xi}}{_{\text{l,}i}}}}\text{sa}{\text{t}}\left(k_{1}\mathbf{x},{v_{\text{m},i}}\right)^{\text{T}}\text{d}\mathbf{x}\label{Vli}
\end{equation}
where $k_{1}$ is an adjustable parameter, $i=1,2,\cdots,M$. From
the definition, $V_{\text{l},i}\geq0.$ According to Thomas' Calculus
\cite[p. 911]{Thomas(2009)}, one has 
\begin{equation}
V_{\text{l},i}=\int_{0}^{t}\text{sat}\left(k_{1}\boldsymbol{\tilde{\xi}}{}_{\text{l,}i}\left(\tau\right),v_{\text{m},i}\right)^{\text{T}}\boldsymbol{\dot{\tilde{\xi}}}{}_{\text{l,}i}\left(\tau\right)\text{d}\tau\mathbf{.}\label{Vli1}
\end{equation}
The objective of the designed velocity command is to make $V_{\text{l},i}$
be zero. This implies that $\left\Vert \boldsymbol{\tilde{\xi}}{_{\text{l,}i}}\right\Vert $
goes down to zero according to the property (\ref{Vli}), namely the
$i$th UAV approaches the finishing line $\overline{{{\mathbf{p}}_{\text{t,2}}{\mathbf{p}}_{\text{t,3}}}}$.

\subsubsection{Barrier Function for Avoiding Conflict with Other UAVs}

\begin{figure}[h]
\begin{centering}
\includegraphics{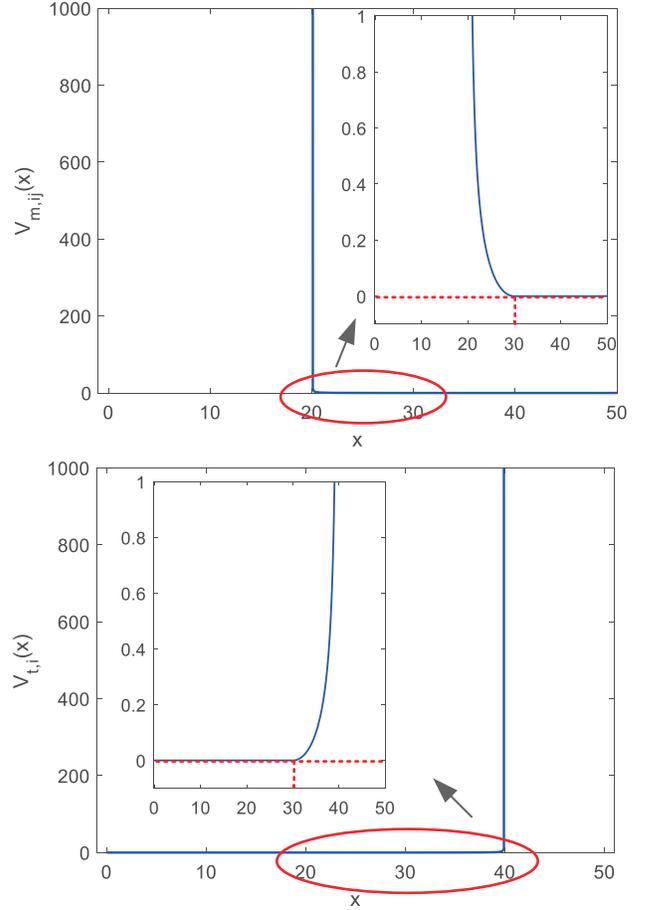} 
\par\end{centering}
\caption{Barrier functions for avoiding collision and keeping within virtual
tube.}
\label{Lyapunovfun} 
\end{figure}

Define 
\begin{equation}
V_{\text{m},ij}=\frac{k_{2}\sigma_{\text{m}}\left(\left\Vert \boldsymbol{\tilde{\xi}}{}_{\text{m,}ij}\right\Vert \right)}{\left(1+\epsilon_{\text{m}}\right)\left\Vert \boldsymbol{\tilde{\xi}}{}_{\text{m,}ij}\right\Vert -2r_{\text{s}}s\left(\frac{\left\Vert \boldsymbol{\tilde{\xi}}{}_{\text{m,}ij}\right\Vert }{2r_{\text{s}}},\epsilon_{\text{s}}\right)}.\label{Vmij}
\end{equation}
Here $\sigma_{\text{m}}\left(x\right)\triangleq\sigma\left(x,2r_{\text{s}},r_{\text{a}}+r_{\text{s}}\right)$,
where $\sigma\left(\cdot\right)$ is defined in (\ref{zerofunction}).\ When
$r_{\text{s}}=10,$ $r_{\text{a}}=20,$ $\epsilon_{\text{m}}=10^{-6},$
${{k}_{2}=1,}$ the function $V_{\text{m},ij}$ is shown in Figure
\ref{Lyapunovfun} (upper plot), where $V_{\text{m},ij}\left(x\right)=0$
as $x\geq r_{\text{s}}+r_{\text{a}}=30$ and $V_{\text{m},ij}\left(x\right)$
is increased sharply as $x\rightarrow0$ from $x=30.$ The function
$V_{\text{m},ij}$ has the following properties: 
\begin{itemize}
\item Property (i). $\partial V_{\text{m},ij}\left/\partial\left\Vert \boldsymbol{\tilde{\xi}}{}_{\text{m,}ij}\right\Vert \right.\leq0$
as $V_{\text{m},ij}\ $is a nonincreasing function with respect to
$\left\Vert \boldsymbol{\tilde{\xi}}{_{\text{m,}ij}}\right\Vert $; 
\item Property (ii). If $\left\Vert \boldsymbol{\tilde{\xi}}{}_{\text{m,}ij}\right\Vert >r_{\text{a}}+r_{\text{s}}{,}$
namely $\mathcal{A}_{i}\cap\mathcal{S}_{j}=\varnothing$ and $\mathcal{A}_{j}\cap\mathcal{S}_{i}=\varnothing,$
then $V_{\text{m},ij}=0$ and $\partial V_{\text{m},ij}\left/\partial\left\Vert \boldsymbol{\tilde{\xi}}{_{\text{m,}ij}}\right\Vert \right.=0$;
if $V_{\text{m},ij}=0,$ then $\left\Vert \boldsymbol{\tilde{\xi}}{_{\text{m,}ij}}\right\Vert >r_{\text{a}}+r_{\text{s}}>2r_{\text{s}};$ 
\item Property (iii). If $0<\left\Vert \boldsymbol{\tilde{\xi}}{}_{\text{m,}ij}\right\Vert <2r_{\text{s}}{,}$
namely $\mathcal{S}_{j}\cap\mathcal{S}_{i}\neq\varnothing$ (they
may not collide in practice), then there exists a sufficiently small
$\epsilon_{\text{s}}>0$ such that 
\begin{equation}
V_{\text{m},ij}\approx\frac{k_{2}}{\epsilon_{\text{m}}\left\Vert \boldsymbol{\tilde{\xi}}{}_{\text{m,}ij}\right\Vert }\geq\frac{k_{2}}{2\epsilon_{\text{m}}r_{\text{s}}}.\label{Vmijd}
\end{equation}
\end{itemize}
The objective of the designed velocity command is to make $V_{\text{m},ij}$
be zero or as small as possible. According to property (ii), this
implies $\left\Vert \boldsymbol{\tilde{\xi}}{_{\text{m,}ij}}\right\Vert >2r_{\text{s}}{,}$
namely the $i$th UAV will not conflict with the $j$th UAV.

\subsubsection{Barrier Function for Keeping within virtual tube}

Define 
\[
V_{\text{t},i}=\frac{k_{3}\sigma_{\text{t}}\left(r_{\text{t}}-\left\Vert \boldsymbol{\tilde{\xi}}{}_{\text{t},i}\right\Vert \right)}{\left(r_{\text{t}}-r_{\text{s}}\right)-\left\Vert \boldsymbol{\tilde{\xi}}{}_{\text{t},i}\right\Vert s\left(\frac{r_{\text{t}}-r_{\text{s}}}{\left\Vert \boldsymbol{\tilde{\xi}}{}_{\text{t},i}\right\Vert +\epsilon_{\text{t}}},\epsilon_{\text{s}}\right)}
\]
where $\sigma_{\text{t}}\left(x\right)\triangleq\sigma\left(x,r_{\text{s}},r_{\text{a}}\right)$.\ When
$r_{\text{t}}=50,$ $r_{\text{s}}=10,$ $r_{\text{a}}=20,$ $\epsilon_{\text{t}}=10^{-6},$
${{k}_{3}=1},$ the function $V_{\text{t},i}\left(x\right)$ is shown
in Figure \ref{Lyapunovfun} (lower plot), where $V_{\text{t},i}\left(x\right)=0$
as $x\leq r_{\text{t}}-r_{\text{a}}=30$ and $V_{\text{t},i}\left(x\right)$
is increased sharply as $x\rightarrow40$ from $x=30.$ The function
$V_{\text{t},i}$ has the following properties:

(i) $\partial V_{\text{t},i}\left/\partial\left\Vert \boldsymbol{\tilde{\xi}}{}_{\text{t},i}\right\Vert \right.\geq0$
as $V_{\text{t},i}\ $is a nondecreasing function with respect to
$\left\Vert \boldsymbol{\tilde{\xi}}{_{\text{t},i}}\right\Vert $;

(ii) if $r_{\text{t}}-\left\Vert \boldsymbol{\tilde{\xi}}{}_{\text{t},i}\right\Vert \geq r_{\text{a}}{,}$
namely the edges of the virtual tube are out of the avoidance area
of the $i$th UAV, then $\sigma_{\text{t}}\left(r_{\text{t}}-\left\Vert \boldsymbol{\tilde{\xi}}{_{\text{t},i}}\right\Vert \right)=0;$
consequently, $V_{\text{t},i}=0$ and $\partial V_{\text{t},i}\left/\partial\left\Vert \boldsymbol{\tilde{\xi}}{_{\text{t},i}}\right\Vert \right.=0$;

(iii) if $r_{\text{t}}-\left\Vert \boldsymbol{\tilde{\xi}}{}_{\text{t},i}\right\Vert <r_{\text{s}},$
namely one edge of the virtual tube has entered into the safety area
of the $i$th UAV, then 
\[
\sigma_{\text{t},i}\left(r_{\text{t}}-\left\Vert \boldsymbol{\tilde{\xi}}{_{\text{t},i}}\right\Vert \right)=1
\]
and there exists a sufficiently small $\epsilon_{\text{s}}>0$ such
that 
\[
s\left(\frac{r_{\text{t}}-r_{\text{s}}}{\left\Vert \boldsymbol{\tilde{\xi}}{}_{\text{t},i}\right\Vert +\epsilon_{\text{t}}},\epsilon_{\text{s}}\right)\approx\frac{r_{\text{t}}-r_{\text{s}}}{\left\Vert \boldsymbol{\tilde{\xi}}{}_{\text{t},i}\right\Vert +\epsilon_{\text{t}}}<1.
\]
As a result, 
\[
V_{\text{t},i}\approx\frac{k_{3}\left(\left\Vert \boldsymbol{\tilde{\xi}}{}_{\text{t},i}\right\Vert +\epsilon_{\text{t}}\right)}{\epsilon_{\text{t}}\left(r_{\text{t}}-r_{\text{s}}\right)}
\]
which will be very large if $\epsilon_{\text{t}}$ is very small.

The objective of the designed velocity command is to make $V_{\text{t},i}$
be zero. This implies $r_{\text{t}}-\left\Vert \boldsymbol{\tilde{\xi}}{_{\text{t},i}}\right\Vert \geq r_{\text{a}}{\ }$according
to property (ii), namely the $i$th UAV will keep within the virtual
tube.

\subsection{Controller Design}

The velocity command is designed as 
\begin{equation}
\mathbf{v}_{\text{c},i}=\mathbf{v}_{\text{T},i}\label{*}
\end{equation}

\begin{align}
\mathbf{v}_{\text{T},i} & =-\text{sat}\Bigg(\underset{\text{Line Approaching}}{\underbrace{\mathbf{A}_{\text{t,23}}\text{sat}\left(k_{1}\boldsymbol{\tilde{\xi}}{}_{\text{l,}i},v_{\text{m},i}\right)}}+\underset{\text{UAV Avoidance}}{\underbrace{\underset{j\in\mathcal{N}_{\text{m},i}}{{\displaystyle \sum}}-b_{ij}\boldsymbol{\tilde{\xi}}_{\text{m,}ij}}}\Bigg.\nonumber \\
 & \Bigg.+\underset{\text{Tunnel Keeping}}{\underbrace{c_{i}\mathbf{A}_{\text{t,12}}\boldsymbol{\tilde{\xi}}_{\text{t},i}}},v_{\text{m},i}\Bigg)\label{control_highway_dis}
\end{align}
with\footnote{$b_{ij}\geq0$ according to the property (i) of $V_{\text{m},ij};$
$c_{i}\geq0$ according to the property (i) of $V_{\text{t},i}.$} 
\begin{align}
b_{ij} & =-\frac{\partial V_{\text{m},ij}}{\partial\left\Vert \boldsymbol{\tilde{\xi}}{}_{\text{m,}ij}\right\Vert }\frac{1}{\left\Vert \boldsymbol{\tilde{\xi}}{}_{\text{m,}ij}\right\Vert }\label{bij}\\
c_{i} & =\frac{\partial V_{\text{t},i}}{\partial\left\Vert \boldsymbol{\tilde{\xi}}{}_{\text{t},i}\right\Vert }\frac{1}{\left\Vert \boldsymbol{\tilde{\xi}}{}_{\text{t},i}\right\Vert }.\label{ci}
\end{align}
This is a distributed control form. Unlike the formation control,
neighboring UAVs' IDs of a UAV are not required. By active detection
devices such as cameras or radars may only detect neighboring UAVs'
position and velocity but no IDs, because these UAVs may look alike.
This implies that the proposed distributed control can work autonomously
without communication.

\textbf{Remark 3}. It is noticed that the velocity command (\ref{control_highway_dis})
is saturated, whose norm will not exceed ${v_{\text{m},i}.}$ If the
case such as $\left\Vert \boldsymbol{\tilde{\xi}}{_{\text{m,}ij_{i}}}\right\Vert <2r_{\text{s}}$
happens in practice due to unpredictable uncertainties out of the
assumptions we make, this may not imply that the $i$th UAV has collided
the $j_{i}$th UAV physically. In this case, the velocity command
(\ref{*}) degenerates to be 
\begin{align*}
\mathbf{v}_{\text{c},i} & =\mathbf{-}\text{sa}{\text{t}}\Bigg(\mathbf{A}_{\text{23}}\text{sa}{\text{t}}\left(k_{1}\boldsymbol{\tilde{\xi}}{_{\text{l,}i}},v_{\text{m},i}\right)-\underset{j=1,j\neq i,j_{i}}{\overset{M}{
{\displaystyle \sum}
}}b_{ij}\boldsymbol{\tilde{\xi}}_{\text{m,}ij}\Bigg.\\
 & \Bigg.+c_{i}\mathbf{A}_{\text{t,12}}\boldsymbol{\tilde{\xi}}_{\text{t},i}-b_{ij_{i}}\boldsymbol{\tilde{\xi}}{_{\text{m,}ij_{i}}},{v_{\text{m},i}}\Bigg)
\end{align*}
with $b_{ij_{i}}\approx\frac{{{k}_{2}}}{\epsilon_{\text{m}}}\frac{1}{\left\Vert \boldsymbol{\tilde{\xi}}{_{\text{m,}ij_{i}}}\right\Vert ^{3}}.$
Since $\epsilon_{\text{m}}$ is chosen to be sufficiently small, the
term $b_{ij_{i}}\boldsymbol{\tilde{\xi}}{_{\text{m,}ij_{i}}}$ will
dominate\footnote{Furthermore, we assume that the $i$th UAV does not conflict with
others except for the $j_{i}$th UAV, or not very close to the edges
of the virtual tube.} so that the velocity command $\mathbf{v}_{\text{c},i}$ becomes 
\[
\mathbf{v}_{\text{c},i}\approx\text{sa}{\text{t}}\left(\frac{{{k}_{2}}}{\epsilon_{\text{m}}}\frac{1}{\left\Vert \boldsymbol{\tilde{\xi}}{_{\text{m,}ij_{i}}}\right\Vert ^{2}}\frac{\boldsymbol{\tilde{\xi}}{_{\text{m,}ij_{i}}}}{\left\Vert \boldsymbol{\tilde{\xi}}{_{\text{m,}ij_{i}}}\right\Vert },{v_{\text{m},i}}\right).
\]
This implies that, by recalling (\ref{mmodel}), $\left\Vert \boldsymbol{\tilde{\xi}}{_{\text{m,}ij_{i}}}\right\Vert $
will be increased very fast so that the $i$th UAV can keep away from
the $j_{i}$th UAV immediately.

\textbf{Remark 4}. In practice, the case such as $r_{\text{h}}-\left\Vert \boldsymbol{\tilde{\xi}}{_{\text{h},i}}\right\Vert <r_{\text{s}}$
may still happen in practice due to unpredictable uncertainties out
of the assumptions we make. In this case, since $\epsilon_{\text{t}}$
is chosen to be sufficiently small, the velocity command $\mathbf{v}_{\text{c},i}$
becomes 
\[
\mathbf{v}_{\text{c},i}\approx\text{sa}{\text{t}}\left(-\frac{{{k}_{3}}}{\epsilon_{\text{t}}\left(r_{\text{t}}-r_{\text{s}}\right)\left\Vert \boldsymbol{\tilde{\xi}}{_{\text{t},i}}\right\Vert }\mathbf{A}_{\text{t,12}}\boldsymbol{\tilde{\xi}}{_{\text{t},i}},{v_{\text{m},i}}\right).
\]
This implies that, by recalling (\ref{tmodel}), $\left\Vert \mathbf{A}_{\text{t,12}}\boldsymbol{\tilde{\xi}}{_{\text{t},i}}\right\Vert $
will be decreased so that the $i$th UAV can return back to the virtual
tube immediately.

\subsection{Stability Analysis}

In order to investigate the basic virtual tube passing problem, a
function is defined as follows 
\[
{V}=\underset{i=1}{\overset{M}{
{\displaystyle \sum}
}}\left(V_{\text{l},i}+\frac{1}{2}\underset{j=1,j\neq i}{\overset{M}{
{\displaystyle \sum}
}}V_{\text{m},ij}+V_{\text{t},i}\right).
\]
The derivative of ${V}$ along the solution to (\ref{lmodel}),(\ref{mmodel}),(\ref{tmodel})
is 
\begin{align*}
{\dot{V}} & =\underset{i=1}{\overset{M}{
{\displaystyle \sum}
}}\Bigg(\text{sa}{\text{t}}\left({{k}_{1}}\boldsymbol{\tilde{\xi}}{_{\text{l,}i}},{v_{\text{m},i}}\right)^{\text{T}}\mathbf{A}_{\text{t,23}}\mathbf{v}_{\text{c},i}\Bigg.\\
 & \Bigg.-\frac{{1}}{2}\underset{j=1,j\neq i}{\overset{M}{
{\displaystyle \sum}
}}b_{ij}\boldsymbol{\tilde{\xi}}_{\text{m,}ij}^{\text{T}}\left(\mathbf{v}_{\text{c},i}-\mathbf{v}_{\text{c},j}\right)+c_{i}\boldsymbol{\tilde{\xi}}_{\text{t},i}^{\text{T}}\mathbf{A}_{\text{t,12}}\mathbf{v}_{\text{c},i}\Bigg)\\
 & =\underset{i=1}{\overset{M}{
{\displaystyle \sum}
}}\Bigg(\mathbf{A}_{\text{t,23}}\text{sa}{\text{t}}\left({{k}_{1}}\boldsymbol{\tilde{\xi}}{_{\text{l,}i}},{v_{\text{m},i}}\right)-\underset{j=1,j\neq i}{\overset{M}{
{\displaystyle \sum}
}}b_{ij}\boldsymbol{\tilde{\xi}}_{\text{m,}ij}\Bigg.\\
 & \Bigg.+c_{i}\mathbf{A}_{\text{t,12}}\boldsymbol{\tilde{\xi}}_{\text{t},i}\Bigg)^{\text{T}}\mathbf{v}_{\text{c},i}.
\end{align*}
Since 
\[
\underset{j=1,j\neq i}{\overset{M}{
{\displaystyle \sum}
}}b_{ij}\boldsymbol{\tilde{\xi}}_{\text{m,}ij}=\underset{j\in\mathcal{N}_{\text{m},i}}{\overset{}{
{\displaystyle \sum}
}}b_{ij}\boldsymbol{\tilde{\xi}}_{\text{m,}ij}
\]
by using the velocity input (\ref{*}), ${\dot{V}}$ becomes 
\begin{equation}
{\dot{V}}\leq0.\label{dV2}
\end{equation}

Before introducing the main result, two lemmas are needed.

\textbf{Lemma 2}. Under \textit{Assumptions 1-4},\textit{ }suppose
that the velocity command is designed as {(\ref{*}). }Then there
exist sufficiently small $\epsilon_{\text{m}},\epsilon_{\text{s}}>0$
in $b_{ij}$ and $\epsilon_{\text{t}}>0$ in $c_{i}$ such that $\left\Vert \boldsymbol{\tilde{\xi}}{_{\text{m,}ij}}\left(t\right)\right\Vert >2r_{\text{s}},$
$\left\Vert \boldsymbol{\tilde{\xi}}{_{\text{t},i}}\right\Vert <r_{\text{t}}-r_{\text{s}},$
$t\in\lbrack0,\infty)$ for all ${{\mathbf{p}}_{i}(0)}$, $i=1,\cdots,M.$

\emph{Proof}. See \emph{Appendix}. $\square$

With \emph{Lemmas 1-2} in hand, we can state the main result.

\textbf{Theorem 1}. Under \textit{Assumptions 1-4},\textit{ }suppose
(i) the velocity command in the distributed form is designed as in
{(\ref{*}); }(ii) given ${\epsilon}_{\text{0}}\in{
\mathbb{R}
}_{+}{,}$ if (\ref{arrivialairway}) is satisfied, then $b_{ij}\equiv0$ and
$c_{i}\equiv0$ (this implies that the $i$th UAV is removed from
the virtual tube mathematically). Then, for given ${\epsilon}_{\text{0}}\in{
\mathbb{R}
}_{+}$, there exist sufficiently small $\epsilon_{\text{m}},r_{\text{s}}\in{\mathbb{R}}_{+}$
in $b_{ij}$, $\epsilon_{\text{t}},\in{\mathbb{R}}_{+}$ in $c_{i}$
and $t_{1}\in{
\mathbb{R}
}_{+}$ such that all UAVs can satisfy (\ref{arrivialairway}) as $t\geq t_{1},$
meanwhile $\left\Vert \boldsymbol{\tilde{\xi}}{_{\text{m,}ij}}\left(t\right)\right\Vert >2r_{\text{s}},$
$\left\Vert \boldsymbol{\tilde{\xi}}{_{\text{t},i}}\right\Vert <r_{\text{t}}-r_{\text{s}},$
$t\in\lbrack0,\infty)$ for all ${{\mathbf{p}}_{i}(0)}$, $i=1,\cdots,M.$

\textit{Proof}. According to \textit{Lemma 2}, these VTOL UAVs are
able to avoid conflict with each other and keep within the virtual
tube, namely $\left\Vert \boldsymbol{\tilde{\xi}}{_{\text{m,}ij}}\left(t\right)\right\Vert >2r_{\text{s}}$,
$\left\Vert \mathbf{\tilde{\mathbf{p}}}{_{\text{t},i}}\right\Vert <r_{\text{t}},$
$i\neq j,$ $i,j=1,2,\cdots,M.$ In the following, the reason why
the $i$th UAV is able to approach the finishing line $\overline{{{\mathbf{p}}_{\text{t,2}}{\mathbf{p}}_{\text{t,3}}}}$
is given.

The function ${V}$ is not a Lyapunov function. The \textit{invariant
set theorem\cite[p. 69]{Slotine(1991)} }is used to do the analysis\textit{.} 
\begin{itemize}
\item First, we will study the property of function $V$. Let $\Omega=\left\{ \left.\boldsymbol{\xi}_{1},\cdots\boldsymbol{\xi}_{M}\right\vert {V}\left(\boldsymbol{\xi}_{1},\cdots\boldsymbol{\xi}_{M}\right)\leq l\right\} ,$
$l>0.$ According to \textit{Lemma 2}, $V_{\text{m},ij},$ $V_{\text{t},i}>0.$
Therefore, ${V}\left(\boldsymbol{\xi}_{1},\cdots\boldsymbol{\xi}_{M}\right)\leq l$
implies $\underset{i=1}{\overset{M}{
{\displaystyle \sum}
}}V_{\text{l},i}\leq l.$ Furthermore, according to \textit{Lemma 1(iii)}, $\Omega$ is bounded.
When $\left\Vert \left[\begin{array}{ccc}
\boldsymbol{\xi}_{1} & \cdots & \boldsymbol{\xi}_{M}\end{array}\right]\right\Vert \rightarrow\infty,$ then $\underset{i=1}{\overset{M}{
{\displaystyle \sum}
}}V_{\text{l},i}\rightarrow\infty$ according to \textit{Lemma 1(ii)}, namley ${V}\rightarrow\infty$.
Therefore the function $V$ satisfies the condition that the invariant
set theorem is requires. 
\item Secondly, we will find the largest invariant set, then show all UAVs
can pass the finishing line. Now, recalling the property (\ref{saturation1}),
${\dot{V}}={0}$ if and only if 
\begin{equation}
\mathbf{A}_{\text{t,23}}\text{sa}{\text{t}}\left({{k}_{1}}\boldsymbol{\tilde{\xi}}{_{\text{l,}i}},{v_{\text{m},i}}\right)-\underset{j=1,j\neq i}{\overset{M}{
{\displaystyle \sum}
}}b_{ij}\boldsymbol{\tilde{\xi}}_{\text{m,}ij}+c_{i}\mathbf{A}_{\text{t,12}}\boldsymbol{\tilde{\xi}}_{\text{t},i}=\mathbf{0}\label{equilibriumTh5_v}
\end{equation}
where $i=1,\cdots,M$. Then $\mathbf{v}_{\text{c},i}=\mathbf{0\ }$according
to ({\ref{*}}). Consequently, by (\ref{positionmodel_ab_con_i}),
the system cannot get ``stuck''\ at an equilibrium value other
than $\mathbf{v}_{i}=\mathbf{0}$. The equation (\ref{equilibriumTh5_v})
can be further written as 
\begin{equation}
{{k}_{1}{\kappa}_{v_{\text{m},i}}}\mathbf{A}_{\text{t,23}}{{\mathbf{\tilde{p}}}_{\text{l,}i}}-\underset{j=1,j\neq i}{\overset{M}{
{\displaystyle \sum}
}}b_{ij}\mathbf{\tilde{p}}_{\text{m,}ij}+c_{i}\mathbf{A}_{\text{t,12}}\mathbf{\tilde{\mathbf{p}}}_{\text{t},i}=\mathbf{0}.\label{equilibriumTh5}
\end{equation}
Let the $1$st UAV be ahead, the closest to the finishing line $\overline{{{\mathbf{p}}_{\text{t,2}}{\mathbf{p}}_{\text{t,3}}}}$.
Let us examine the following equation related to the 1st UAV that
\begin{equation}
{{k}_{1}{\kappa}_{v_{\text{w},1}}}\mathbf{A}_{\text{t,23}}{{\mathbf{\tilde{p}}}_{\text{l,}1}}-\underset{j=2}{\overset{M}{
{\displaystyle \sum}
}}b_{1j}\mathbf{\tilde{p}}_{\text{m,}1j}+c_{1}\mathbf{A}_{\text{t,12}}\mathbf{\tilde{\mathbf{p}}}_{\text{t},1}=\mathbf{0}.\label{equilibriumTh5_1st}
\end{equation}
Since the $1$st UAV is ahead, we have 
\begin{equation}
\left({{\mathbf{p}}_{\text{t,2}}}-{{\mathbf{p}}_{\text{t,1}}}\right)^{\text{T}}\mathbf{\tilde{p}}_{\text{m,}1j}\geq0\label{1st}
\end{equation}
where ``$=$''\ hold if and only if the $j$th UAV is as ahead
as the $1$st one. On the other hand, one has 
\begin{equation}
\left({{\mathbf{p}}_{\text{t,2}}}-{{\mathbf{p}}_{\text{t,1}}}\right)^{\text{T}}\mathbf{A}_{\text{t,12}}=0.\label{perpendicular}
\end{equation}
Multiplying the term $\left({{\mathbf{p}}_{\text{t,2}}}-{{\mathbf{p}}_{\text{t,1}}}\right){^{\text{T}}}$
at the left side of {(\ref{equilibriumTh5_1st}) results in} 
\[
{{k}_{1}{\kappa}_{v_{\text{w},1}}\left({{\mathbf{p}}_{\text{t,2}}}-{{\mathbf{p}}_{\text{t,1}}}\right){^{\text{T}}}}\mathbf{A}_{\text{t,23}}{{\mathbf{\tilde{p}}}_{\text{l,}1}}=\left({{\mathbf{p}}_{\text{t,2}}}-{{\mathbf{p}}_{\text{t,1}}}\right){^{\text{T}}}\underset{j=2}{\overset{M}{
{\displaystyle \sum}
}}b_{1j}\mathbf{\tilde{p}}_{\text{m,}1j}
\]
where {(\ref{perpendicular}) is used. Since }$b_{1j}\geq0$ and
{(\ref{1st}) holds for the 1st UAV, one has} 
\[
{\left({{\mathbf{p}}_{\text{t,2}}}-{{\mathbf{p}}_{\text{t,1}}}\right){^{\text{T}}}}\mathbf{A}_{\text{t,23}}{{\mathbf{\tilde{p}}}_{\text{l,}1}\geq0.}
\]
Since ${\left({{\mathbf{p}}_{\text{t,2}}}-{{\mathbf{p}}_{\text{t,1}}}\right){^{\text{T}}}}\mathbf{A}_{\text{t,23}}{{\mathbf{\tilde{p}}}_{\text{l,}1}}\left(0\right)<{0\ }$according
to \textit{Assumption 2}, owing to the continuity, given ${\epsilon}_{\text{0}}\in
\mathbb{R}
_{+},$ there must exist a time $t_{11}\in
\mathbb{R}
_{+}$ such that 
\[
\left(\frac{{{\mathbf{p}}_{\text{t,2}}}-{{\mathbf{p}}_{\text{t,1}}}}{\left\Vert {{\mathbf{p}}_{\text{t,2}}}-{{\mathbf{p}}_{\text{t,1}}}\right\Vert }\right)^{\text{T}}\left(\mathbf{p}_{i}\left(t\right)-{{\mathbf{p}}_{\text{t,2}}}\right)\geq-{\epsilon}_{\text{0}}
\]
as $t\geq t_{11}.$ At the time $t_{11},$ the $1$st UAV is removed
from {(\ref{equilibriumTh5}) according to \textit{Assumption 4},
namely it quits }the virtual tube. The left problem is to consider
the $M-1$ UAVs, namely $2$nd, $3$rd, ..., $M$th UAVs. We can repeat
the analysis above to conclude this proof. $\square$ 
\end{itemize}

\section{Controller Design for General virtual tube Passing Problem}

So far, we have solved the basic\textbf{ }virtual tube passing problem.
Then, we are going to solve the general virtual tube passing problem.
First, we define different areas for the whole airspace. Then, the
general virtual tube passing problem is decomposed into several basic\textbf{
}virtual tube passing problems. As a result, for UAVs in different
areas, they have different controllers, like ({\ref{*}}). Combining
them together, the final controller is obtained.

\subsection{Area Definition}

\begin{figure}[h]
\begin{centering}
\includegraphics{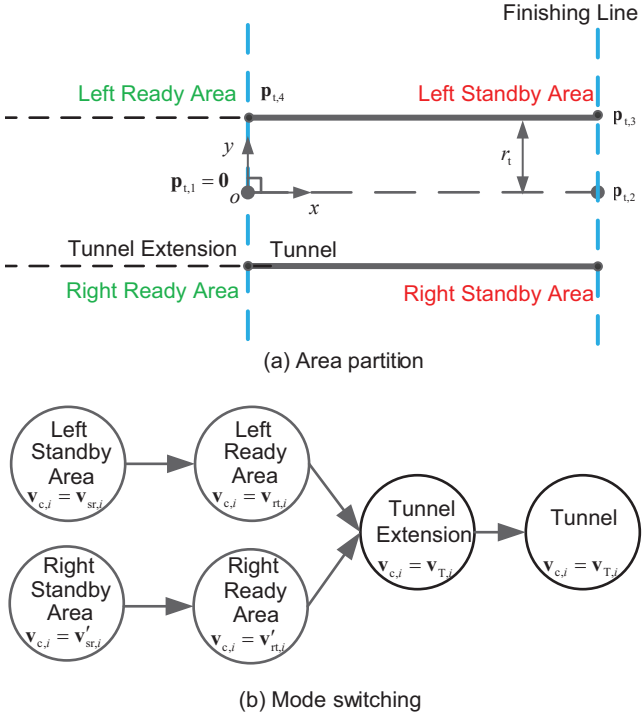} 
\par\end{centering}
\caption{Area definition and flight sequence.}
\label{modeswitching} 
\end{figure}

As shown in Figure \ref{modeswitching}(a), the whole airspace is
divided into six areas, namely \emph{Left} \emph{Standby Area}, \emph{Left}
\emph{Ready Area}, \emph{Right} \emph{Standby Area}, \emph{Right}
\emph{Ready Area}, \emph{virtual tube, }and\emph{ virtual tube Extension}.
Moreover, the Earth-fixed coordinate frame is built. For simplicity,
let ${{\mathbf{p}}_{\text{t,1}}}=\mathbf{0\ }$with $x$-axis pointing
to ${{\mathbf{p}}_{\text{t,2}}}$ and $y$-axis pointing to its left
side. 
\begin{itemize}
\item \emph{Left} \emph{Standby Area }and\emph{ Right} \emph{Standby Area
}are the areas on the outside of \emph{virtual tube} and the right
side of \emph{Starting Line} $\overline{{{\mathbf{p}}_{\text{t,1}}{\mathbf{p}}_{\text{t,4}}}},$
where ${{\mathbf{p}}_{\text{t,4}}=[0}$ $r_{\text{t}}{]}^{\text{T}}.\mathcal{\ }$Concretely,
if 
\[
\boldsymbol{\xi}_{i}\left(1\right)>0,\boldsymbol{\xi}_{i}\left(2\right)>r_{\text{t}}
\]
then $\boldsymbol{\xi}_{i}$ is in \emph{Left} \emph{Standby Area.}
If 
\[
\boldsymbol{\xi}_{i}\left(1\right)>0,\boldsymbol{\xi}_{i}\left(2\right)<-r_{\text{t}}
\]
then $\boldsymbol{\xi}_{i}$ is in \emph{Right} \emph{Standby Area.} 
\item \emph{Left Ready Area }and\emph{ Right} \emph{Ready Area }are the
areas on the outside of \emph{virtual tube} \emph{Extension}\ and
the left side of \emph{Starting Line} $\overline{{{\mathbf{p}}_{\text{t,1}}{\mathbf{p}}_{\text{t,4}}}}$.
Concretely, if 
\[
\boldsymbol{\xi}_{i}\left(1\right)\leq0,\boldsymbol{\xi}_{i}\left(2\right)>r_{\text{t}}
\]
then $\boldsymbol{\xi}_{i}$ is in \emph{Left} \emph{Standby Area.}
If 
\[
\boldsymbol{\xi}_{i}\left(1\right)\leq0,\boldsymbol{\xi}_{i}\left(2\right)<-r_{\text{t}}
\]
then $\boldsymbol{\xi}_{i}$ is in \emph{Right} \emph{Standby Area.} 
\item \emph{virtual tube }and\emph{ virtual tube Extension }are\emph{ }a
band\emph{.} Concretely, if 
\begin{align*}
\boldsymbol{\xi}_{i}\left(1\right) & \leq0\text{ \& }\boldsymbol{\xi}_{i}\left(1\right)>\left\Vert \mathbf{p}_{\text{t,1}}-\mathbf{p}_{\text{t,2}}\right\Vert \\
-r_{\text{t}} & \leq\boldsymbol{\xi}_{i}\left(2\right)\leq r_{\text{t}}
\end{align*}
then $\boldsymbol{\xi}_{i}$ is in \emph{virtual tube Extension.}
If 
\begin{align*}
0 & <\boldsymbol{\xi}_{i}\left(1\right)\leq\left\Vert \mathbf{p}_{\text{t,1}}-\mathbf{p}_{\text{t,2}}\right\Vert \\
-r_{\text{t}} & \leq\boldsymbol{\xi}_{i}\left(2\right)\leq r_{\text{t}}
\end{align*}
then $\boldsymbol{\xi}_{i}$ is in \emph{virtual tube.} 
\end{itemize}

\subsection{virtual tube Passing Scheme and Requirements}

As Assumption 2$^{\prime}$ points, at the beginning, UAVs may locate
in the six areas\emph{. }A flight sequence is given shown in Figure
\ref{modeswitching}(b)\emph{. }The requirement is as follows. 
\begin{itemize}
\item From \emph{Left/Right Standby Area} to \emph{Left/Right} \emph{Ready
Area. }{{UAVs}} are required to fly into \emph{Left/Right} \emph{Ready
Area}, meanwhile avoiding conflict with other UAVs and keeping away
from \emph{virtual tube }and\emph{ virtual tube Extension.} 
\item From \emph{Left/Right Ready Area }to\emph{ virtual tube Extension.
}{{UAVs}} are required to fly into \emph{virtual tube Extension},
meanwhile avoiding conflict with other UAVs and keeping away from
\emph{virtual tube.} 
\item From \emph{virtual tube and virtual tube Extension }to \emph{Finishing
Line }$\overline{{{\mathbf{p}}_{\text{t,2}}{\mathbf{p}}_{\text{t,3}}}}.$
{{UAVs}} are required to pass the virtual tube until it arrives
near the finishing line $\overline{{{\mathbf{p}}_{\text{t,2}}{\mathbf{p}}_{\text{t,3}}}}$,
meanwhile avoiding conflict other UAVs and keeping within the virtual
tube and its extension. 
\end{itemize}

\subsection{Controller Design}

\subsubsection{From Standby Area to Ready Area}

\begin{figure}[h]
\begin{centering}
\includegraphics{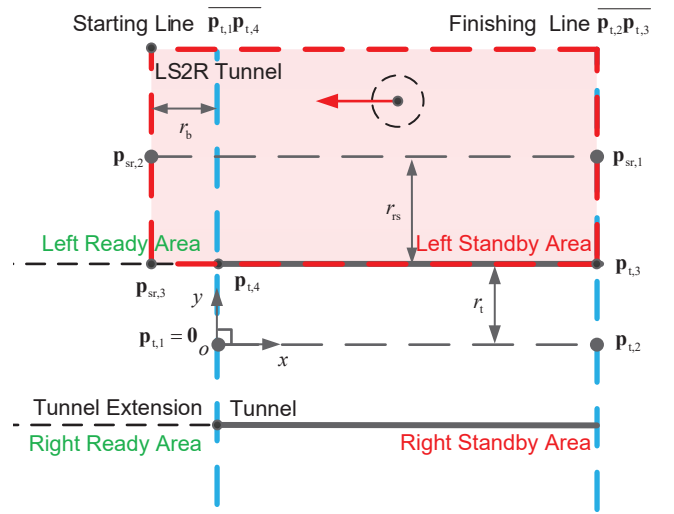} 
\par\end{centering}
\caption{LS2R virtual tube is designed for\emph{ }from Left Standby Area to
Left Ready Area.}
\label{LS2R} 
\end{figure}

As shown in Figure \ref{LS2R}, a virtual virtual tube, named \emph{LS2R
virtual tube}, is designed with the width $2r_{\text{sr}}$ and centerline
starting from ${\mathbf{p}}_{\text{sr,1}}\in{{\mathbb{R}}^{2}}$ to
${\mathbf{p}}_{\text{sr,2}}\in{{\mathbb{R}}^{2},}$ where $r_{\text{sr}}>0$
is often sufficiently large ($10000$ in the following simulation
for example) that takes all UAVs in \emph{Left Standby Area }in the\emph{
}virtual virtual tube. Moreover, we let all UAVs in \emph{Left Standby
Area }approach the finishing line $\overline{\mathbf{p}_{\text{sr,2}}\mathbf{p}_{\text{sr,3}}}.$
Here 
\[
\mathbf{p}_{\text{sr,1}}=\left[\begin{array}{c}
\mathbf{p}_{\text{t,2}}\left(1\right)\\
r_{\text{t}}+r_{\text{sr}}
\end{array}\right],\mathbf{p}_{\text{sr,2}}=\left[\begin{array}{c}
-r_{\text{b}}\\
r_{\text{t}}+r_{\text{sr}}
\end{array}\right],\mathbf{p}_{\text{sr,3}}=\left[\begin{array}{c}
-r_{\text{b}}\\
r_{\text{t}}
\end{array}\right]
\]
where $r_{\text{b}}>0,$ $r_{\text{b}}=r_{\text{a}}\ $for example.
The intersection of \emph{LS2R virtual tube }and\emph{ Left Ready
Area }is a buffer with length $r_{\text{b}}$, which can make a UAV
fly into \emph{Left Ready Area }not only approaching it.\emph{ }According
to (\ref{arrivialairway}), the controller is designed as $\mathbf{v}_{\text{c},i}=\mathbf{v}_{\text{sr},i},$
where 
\begin{align}
\mathbf{v}_{\text{sr},i} & ={\text{sa}{\text{t}}}\big(\mathbf{v}_{\text{l},i}\left({k}_{1},{\mathbf{p}}_{\text{sr,2}},{\mathbf{p}}_{\text{sr,3}}\right)+\mathbf{v}_{\text{m},i}\left({{k}_{2}}\right)\big.\nonumber \\
 & \big.+\mathbf{v}_{\text{t},i}\left({{k}_{3},r_{\text{sr}},{\mathbf{p}}}_{\text{sr,1}},{\mathbf{p}}_{\text{sr,2}}\right),{v_{\text{m},i}}\big)\label{vsrl}
\end{align}
where 
\begin{align}
\mathbf{v}_{\text{l},i}\left({{k}_{1},{\mathbf{p}}_{\text{t,2}}},{{\mathbf{p}}_{\text{t,3}}}\right) & \triangleq-\mathbf{A}_{\text{t,23}}\left({{\mathbf{p}}_{\text{t,2}}},{{\mathbf{p}}_{\text{t,3}}}\right)\text{sa}{\text{t}}\left({{k}_{1}}\boldsymbol{\tilde{\xi}}{_{\text{l,}i}},{v_{\text{m},i}}\right)\label{control_line}\\
\mathbf{v}_{\text{m},i}\left({{k}_{2}}\right) & \triangleq\underset{j\in\mathcal{N}_{\text{m},i}}{\overset{}{
{\displaystyle \sum}
}}b_{ij}\left({{k}_{2}}\right)\boldsymbol{\tilde{\xi}}_{\text{m,}ij}\label{control_mul}\\
\mathbf{v}_{\text{t},i}\left({{k}_{3},r_{\text{t}},{\mathbf{p}}_{\text{t,1}}},{{\mathbf{p}}_{\text{t,2}}}\right) & \triangleq-c_{i}\left({{k}_{3},r_{\text{t}}}\right)\mathbf{A}_{\text{t,12}}\left({{\mathbf{p}}_{\text{t,1}}},{{\mathbf{p}}_{\text{t,2}}}\right)\boldsymbol{\tilde{\xi}}_{\text{t},i}.\label{control_tun}
\end{align}
Furthermore, according to \textit{Theorem 1},\textit{ }UAVs in \emph{Left
Standby Area }will fly into \emph{Left} \emph{Ready Area}, meaning
while avoiding colliding other UAVs and keeping within \emph{LS2R
virtual tube}, namely keeping away\emph{ }from \emph{virtual tube
}and\emph{ virtual tube Extension.}

Similarly, the controller for from \emph{Right Standby Area} to \emph{Right
Ready Area}\ is designed as $\mathbf{v}_{\text{c},i}=\mathbf{v}_{\text{sr},i}^{\prime},$
where 
\begin{align}
\mathbf{v}_{\text{sr},i}^{\prime} & ={\text{sa}{\text{t}}}\big(\mathbf{v}_{\text{l},i}\left({{k}_{1},{\mathbf{p}}}_{\text{sr,2}}^{\prime},{\mathbf{p}}_{\text{sr,3}}^{\prime}\right)+\mathbf{v}_{\text{m},i}\left({{k}_{2}}\right)\big.\nonumber \\
 & \big.+\mathbf{v}_{\text{t},i}\left({{k}_{3},r_{\text{sr}},{\mathbf{p}}}_{\text{sr,1}}^{\prime},{\mathbf{p}}_{\text{sr,2}}^{\prime}\right),{v_{\text{m},i}}\big)\label{vsrr}
\end{align}
with 
\[
\mathbf{p}_{\text{sr,1}}^{\prime}=\left[\begin{array}{c}
\mathbf{p}_{\text{t,2}}\left(1\right)\\
-r_{\text{t}}-r_{\text{sr}}
\end{array}\right],\mathbf{p}_{\text{sr,2}}^{\prime}=\left[\begin{array}{c}
-r_{\text{b}}\\
-r_{\text{t}}-r_{\text{sr}}
\end{array}\right],
\]

\[
\mathbf{p}_{\text{sr,3}}^{\prime}=\left[\begin{array}{c}
-r_{\text{b}}\\
-r_{\text{t}}
\end{array}\right].
\]

\subsubsection{From Ready Area to virtual tube Extension}

\begin{figure}[h]
\begin{centering}
\includegraphics{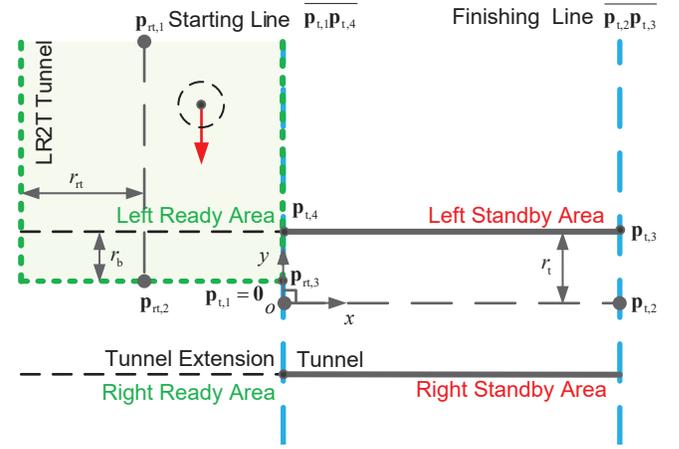} 
\par\end{centering}
\caption{LR2T virtual tube is designed for\emph{ }from Left Ready Area to virtual
tube Extension.}
\label{LR2T} 
\end{figure}

As shown in Figure \ref{LR2T}, a virtual virtual tube, named \emph{LR2T
virtual tube}, is designed with the width $2r_{\text{rt}}$ and centerline
starting from ${\mathbf{p}}_{\text{rt,1}}\in{{\mathbb{R}}^{2}}$ to
${\mathbf{p}}_{\text{rt,2}}\in{{\mathbb{R}}^{2},}$ where $r_{\text{rt}}>0$
is sufficiently large that takes all UAVs in \emph{Left Ready Area
}in the\emph{ }virtual virtual tube. Moreover, all UAVs in \emph{Left
Ready Area }approach the finishing line $\overline{\mathbf{p}_{\text{rt,2}}\mathbf{p}_{\text{rt,3}}}.$
Here 
\[
\mathbf{p}_{\text{rt,1}}=\left[\begin{array}{c}
-r_{\text{rt}}\\
r_{\text{t}}+r_{\text{rt}}
\end{array}\right],\mathbf{p}_{\text{rt,2}}=\left[\begin{array}{c}
-r_{\text{rt}}\\
r_{\text{t}}-r_{\text{b}}
\end{array}\right],\mathbf{p}_{\text{rt,3}}=\left[\begin{array}{c}
0\\
r_{\text{t}}-r_{\text{b}}
\end{array}\right].
\]
The intersection of \emph{LR2T virtual tube }and\emph{ virtual tube
Extension }is a buffer with length $r_{\text{b}}$, which can make
a UAV fly into \emph{virtual tube Extension }not only approaching
it. According to (\ref{arrivialairway}), the controller is designed
as $\mathbf{v}_{\text{c},i}=\mathbf{v}_{\text{rt},i},$ where 
\begin{align}
\mathbf{v}_{\text{rt},i} & ={\text{sa}{\text{t}}}\big(\mathbf{v}_{\text{l},i}\left({{k}_{1},{\mathbf{p}}_{\text{rt,2}},{\mathbf{p}}_{\text{rt,3}}}\right)+\mathbf{v}_{\text{m},i}\left({{k}_{2}}\right)\big.\nonumber \\
 & \big.+\mathbf{v}_{\text{t},i}\left({{k}_{3},r_{\text{rt}},{\mathbf{p}}_{\text{rt,1}},{\mathbf{p}}_{\text{rt,2}}}\right),{v_{\text{m},i}}\big)\label{vrtl}
\end{align}
According to \textit{Theorem 1},\textit{ }UAVs in \emph{Left Ready
Area }will fly into \emph{virtual tube Extension}, meanwhile avoiding
conflict with other UAVs and keeping within \emph{LR2T virtual tube},
namely keeping away\emph{ }from \emph{virtual tube.} Similarly, the
controller for \emph{Right Ready Area} to \emph{virtual tube Extension}\ is
designed as $\mathbf{v}_{\text{c},i}=\mathbf{v}_{\text{rt},i},$ where
\begin{align}
\mathbf{v}_{\text{rt},i}^{\prime} & ={\text{sa}{\text{t}}}\big(\mathbf{v}_{\text{l},i}\left({{k}_{1},{\mathbf{p}}_{\text{rt,2}}^{\prime},{\mathbf{p}}_{\text{rt,3}}^{\prime}}\right)+\mathbf{v}_{\text{m},i}\left({{k}_{2}}\right)\big.\nonumber \\
 & \big.+\mathbf{v}_{\text{t},i}\left({{k}_{3},r_{\text{rt}},{\mathbf{p}}_{\text{rt,1}}^{\prime},{\mathbf{p}}_{\text{rt,2}}^{\prime}}\right),{v_{\text{m},i}}\big)\label{vrtr}
\end{align}
with 
\[
\mathbf{p}_{\text{rt,1}}^{\prime}=\left[\begin{array}{c}
-r_{\text{rt}}\\
-r_{\text{t}}-r_{\text{rt}}
\end{array}\right],\mathbf{p}_{\text{rt,2}}^{\prime}=\left[\begin{array}{c}
-r_{\text{rt}}\\
-r_{\text{t}}+r_{\text{b}}
\end{array}\right],\mathbf{p}_{\text{rt,3}}^{\prime}=\left[\begin{array}{c}
0\\
-r_{\text{t}}+r_{\text{b}}
\end{array}\right].
\]

\subsubsection{Final Controller}

With the design above, the final controller is designed as 
\begin{equation}
\mathbf{v}_{\text{c},i}=\left\{ \begin{array}{ll}
\mathbf{v}_{\text{T},i}\text{ ({\ref{control_highway_dis}})} & \text{if }\boldsymbol{\xi}_{i}\text{ in }\emph{Tunnel\ and\ Tunnel\ Extension}\\
\mathbf{v}_{\text{sr},i}\text{ (\ref{vsrl})} & \text{if }\boldsymbol{\xi}_{i}\text{ in }\emph{Left\ Standby\ Area}\\
\mathbf{v}_{\text{sr},i}^{\prime}\text{ (\ref{vsrr})} & \text{if }\boldsymbol{\xi}_{i}\text{ in }\emph{Right\ Standby\ Area}\\
\mathbf{v}_{\text{rt},i}\text{ (\ref{vrtl})} & \text{if }\boldsymbol{\xi}_{i}\text{ in }\emph{Left\ Ready\ Area}\\
\mathbf{v}_{\text{rt},i}^{\prime}\text{ (\ref{vrtr})} & \text{if }\boldsymbol{\xi}_{i}\text{ in }\emph{Right\ Ready\ Area}
\end{array}\right.\label{control_general}
\end{equation}
where $i=1,2,\cdots,M$. Then, for given ${\epsilon}_{\text{0}}\in{
\mathbb{R}
}_{+}$, there exist sufficiently small $\epsilon_{\text{m}},r_{\text{s}}\in{\mathbb{R}}_{+}$
in $b_{ij}$, $\epsilon_{\text{t}}\in{\mathbb{R}}_{+}$ in $c_{i}$
and $t_{1}\in{
\mathbb{R}
}_{+}$ such that all UAVs can satisfy (\ref{arrivialairway}) as $t\geq t_{1},$
meanwhile $\left\Vert \boldsymbol{\tilde{\xi}}{_{\text{m,}ij}}\right\Vert >2r_{\text{s}}$
and $\left\Vert \boldsymbol{\tilde{\xi}}{_{\text{t,}i}}\right\Vert <r_{\text{t}}-r_{\text{s}}$
when passing the virtual tube, $t\in\lbrack0,\infty)$ for all ${{\mathbf{p}}_{i}(0)}$,
$i=1,\cdots,M.$

\section{Simulation and Experiment}

Simulations and experiments are given in the following to show the
effectiveness of the proposed method, where a video about simulations
and experiments is available on https://youtu.be/LTNG7jWlBdY or http://weibo.ws/nJiiXJ.

\subsection{Simulation}

\subsubsection{Simulation with 40 UAVs}

\begin{figure}[t]
\begin{centering}
\includegraphics[scale=0.8]{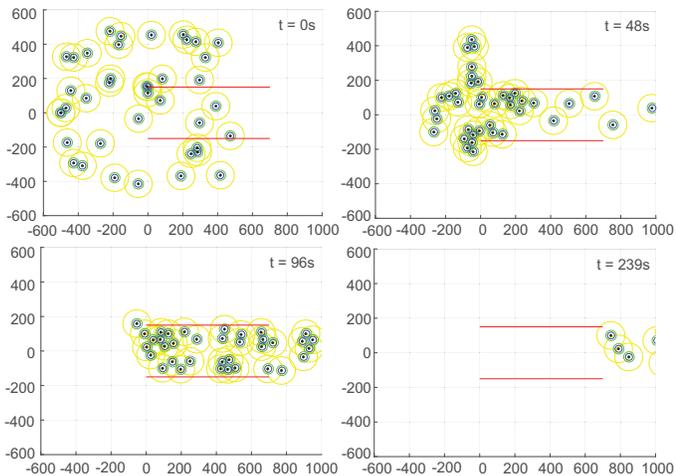} 
\par\end{centering}
\caption{UAVs' positions at different time.}
\label{motion} 
\end{figure}

\begin{figure}[h]
\begin{centering}
\includegraphics{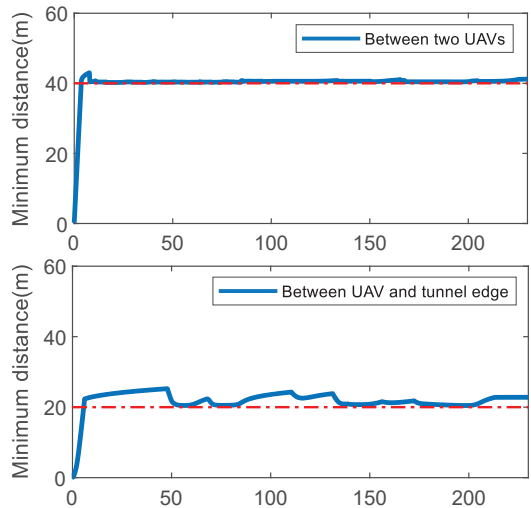} 
\par\end{centering}
\caption{Minimum distance among all UAVs and minimum distance from the virtual
tube edge to all UAVs.}
\label{distance} 
\end{figure}

We consider a scenario of $M=40$ UAVs with $r_{\text{s}}=20$m, $r_{\text{a}}=30$m,
$r_{\text{d}}=80$m, $l_{i}=5$ and $v_{\text{m},i}=5+\frac{i}{4}$m${/}$s,
${i=1,2,\cdots,M}$. The virtual tube is a long horizontal band with
the width $2r_{\text{t}}=300$m and centerline through $\mathbf{p}_{\text{t,1}}=[0\ 0]^{\text{T}}$
m and $\mathbf{p}_{\text{t,2}}=[500\ 0]^{\text{T}}$m. In order {to
show the effectiveness of the proposed method, the }general virtual
tube passing problem is considered directly. {The initial positions
of all UAVs are at everywhere with initial velocities being zero,
shown by} $t=0$s in Figure \ref{motion}, where $\mathbf{p}_{1}\left(0\right)=[0\ 149.9]^{\text{T}}$m
(the 1st UAV has a conflict with the virtual tube); $\mathbf{p}_{2}\left(0\right)=[-500-0.1]^{\text{T}}$m
and ${\mathbf{p}}_{3}\left(0\right)=[-500$ $0.1]^{\text{T}}$m (the
2nd and 3rd UAVs have a conflict with each other initially) are set
intentionally. The parameters of controller are $\epsilon_{\text{m}}=\epsilon_{\text{t}}=\epsilon_{\text{s}}=10^{-6},$
$k_{1}=k_{2}=k_{3}=1$. With these conditions and parameters, the
controller is performed for the simulation getting the results shown
in Figure \ref{motion} and Figure \ref{distance}. As shown in Figure
\ref{motion}, these UAVs enter into the virtual tube according to
the sequence given in Figure \ref{modeswitching}(b), and pass the
finishing line finally about $t=239$s. As shown in Figure \ref{distance}
(upper plot), during this process, the minimum distance of all pairs
of UAVs is $0.2$m because $\left\Vert {\mathbf{p}}_{2}\left(0\right)-{\mathbf{p}}_{3}\left(0\right)\right\Vert =0.2$m.
But, the minimum distance is increased rapidly and then always keeps
above $2r_{\text{s}}=40$m (indicated by a dash line). This shows
the effectiveness of the proposed controller that even if a UAV enters
into the safety area of another UAV, it can keep away from the UAV
rapidly. After this conflict, no conflict between any two UAVs will
happen again. Similarly, as shown in Figure \ref{distance} (lower
plot), during this process, the minimum distance from the virtual
tube is 0.1m because of the 1st UAV being very close to the virtual
tube. But, the minimum distance is increased rapidly and then always
keeps above $r_{\text{s}}=20$m (indicated by a dash line). The reason
can be found in \textit{Remark 4}. After this conflict, no conflict
with the virtual tube will happen again. 

\subsubsection{Comparison of the Calculation Speed of Different Algorithms}

\begin{figure}[ptbh]
\begin{centering}
\includegraphics[scale=0.7]{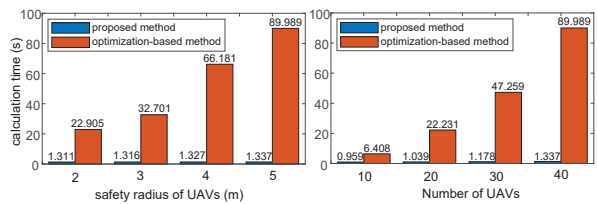} 
\par\end{centering}
\caption{The calculation speed of different algorithms.}
\label{compare} 
\end{figure}

In order to show the advantage of the proposed algorithm, we compare
the calculation speed with an optimization-based algorithm. In \cite{Ingersoll(2016)},
a path-planning algorithm using Bezier curves with the open-source
code at \url{https://github.com/byuflowlab/uav-path-optimization}
is proposed, which can find the optimal solutions to the offline and
online path-planning problem. We design a scenario that contains 10
UAVs with $r_{\text{s}}=5$m, $r_{\text{a}}=7.5$m, $r_{\text{d}}=20$m,
and a virtual tube with the width $2r_{\text{t}}=200$m, centerline
through ${{\mathbf{p}}_{\text{t,1}}}=[0$ $0]^{\text{T}}$m and ${{\mathbf{p}}_{\text{t,2}}}=[100\ 0]^{\text{T}}$m.
The initial position of 1st UAV ${\mathbf{p}}_{1}\left(0\right)={{\mathbf{p}}_{\text{t,1}}}=[0$
$0]^{\text{T}}$, while the other UAVs are distributed inside the
virtual tube randomly with the same velocity $\mathbf{v}_{i}=[1\ 0]^{\text{T}}$m${/}$s,
${i=2,\cdots,M}$. For two different algorithms, we design 10 sets
of random initial positions for the other UAVs, run the simulation
on the same computer and record the \emph{average calculation time}
when the 1st UAV passes the finishing line of the virtual tube. Figure
\ref{compare} shows the performance of the density on calculation
speed by the changing the safety radius and the number of UAVs separately.
As shown in Figure \ref{compare}, for the same airspace, if the number
of UAVs increases or the safety radius of UAVs gets larger, the calculation
speed of optimization-based algorithm will decrease rapidly because
the probability of constraint being triggered is increasing, which
brings more complex calculations. On the contrary, the proposed controller
can better deal with such situation, in other words, be suitable for
dense and complex environment. 

\subsection{Flight Experiments}

\begin{figure}[ptbh]
\begin{centering}
\includegraphics{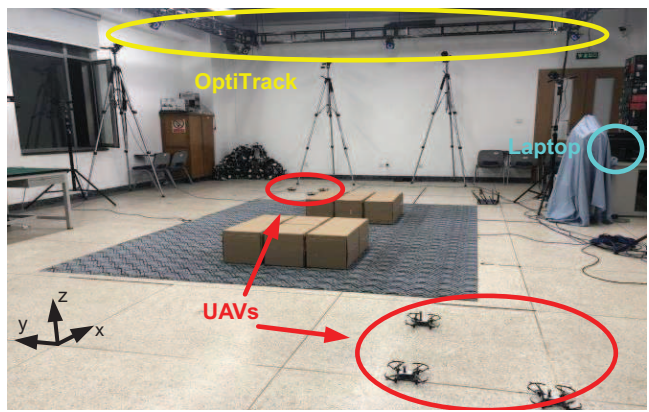} 
\par\end{centering}
\caption{Indoor experiment environment.}
\label{environment} 
\end{figure}

As shown in Fig. \ref{environment}, a set of experiments is carried
out in a laboratory room with an OptiTrack motion capture system installed,
which provides the positions and orientations of UAVs for distributed
control. The laptop is connected to the Tello UAVs and OptiTrack by
a local network, running the proposed controller and a real-time position
plotting module.

\begin{figure}[ptbh]
\begin{centering}
\includegraphics[scale=0.5]{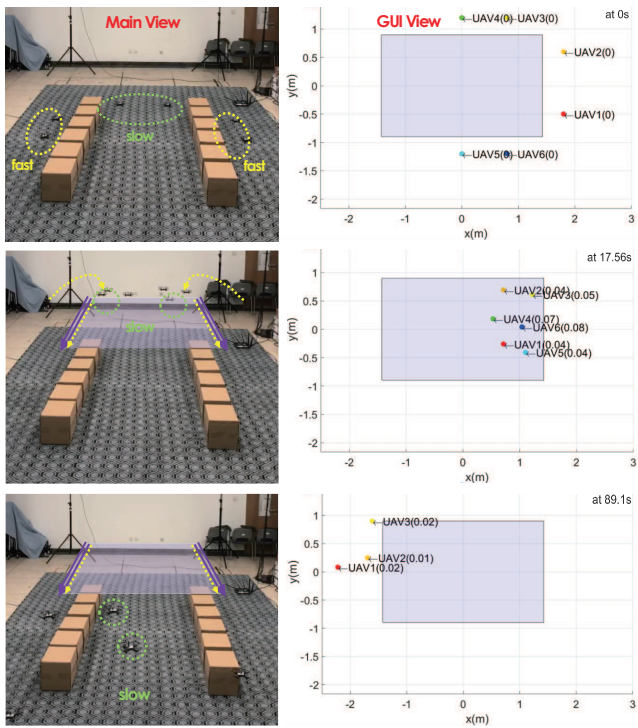} 
\par\end{centering}
\caption{The position, velocity and mode of each UAV during the first experiment}
\label{exp1} 
\end{figure}

In the first experiment, there are $M=6$ UAVs to pass a virtual tube
(indicated by a purple rectangle) as shown in Fig. \ref{exp1}. The
virtual tube is a long horizontal band with the width $2r_{\text{t}}=1.7$m
and centerline through ${{\mathbf{p}}_{\text{t,1}}}=[1.4$ $0]^{\text{T}}$m
and ${{\mathbf{p}}_{\text{t,2}}}=[-1.4\ 0]^{\text{T}}$m. Assuming
that $r_{\text{s}}=0.16$m, $r_{\text{a}}=0.4$m, ${v_{\text{m},1}=v_{\text{m},2}=0.1}$m${/}$s
and ${v_{\text{m},i}=0.2}${m}${/}${s, }${i=3,4,\cdots,M}$.
The parameters of controller (\ref{control_general}) are $\epsilon=r_{\text{s}}=10^{-6},$
$k_{1}=k_{2}=k_{3}=1$ and $r_{\text{b}}=r_{\text{a}},$ $r_{\text{sr}}=r_{\text{rt}}=10000$m.
As shown in Fig \ref{exp1}(a), the initial positions of all UAVs
(indicated by the dotted circle) are at everywhere with initial velocities
being zero. We track UAVs from three perspectives: the main view,
local view and the Graphical User Interface(GUI) view. As shown in
Fig. \ref{exp1}(b), after take-off they all enter into the virtual
tube according to the sequence given in Figure \ref{modeswitching}(b).
By comparing with Fig. \ref{exp1}(c), we can get the result that
the slower UAVs are gradually overtaken by the faster ones. Finally,
as shown in Fig. \ref{exp1}(c), all UAVs complete their routes at
213.88 second, and they always keep a safe distance from others. This
means each UAV passes the virtual tube and avoids each other successfully.

\begin{figure}[tp]
\begin{centering}
\includegraphics[scale=0.5]{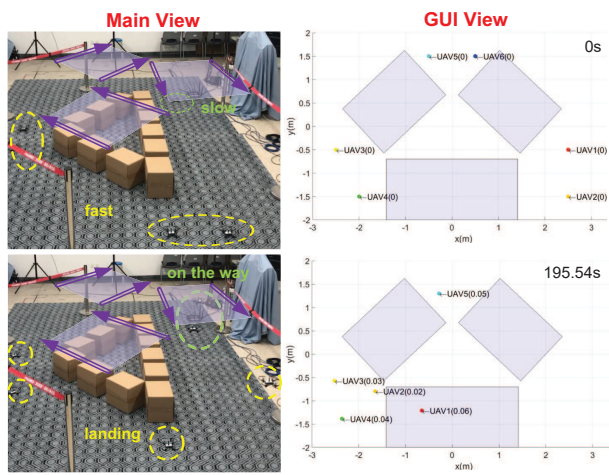} 
\par\end{centering}
\caption{Position and velocity during the experiment.}
\label{exp2} 
\end{figure}

In the experiment, three virtual tubes are added into the flight progress
as shown by $t=0$s in Figure \ref{exp2}. UAVs need to pass the three
virtual tubes in order of clockwise rotation (indicated by the purple
arrow), during which the faster ones will overtake the slow ones.
As shown in Figure \ref{exp2} (lower plot), the faster UAVs pass
the three virtual tube at 195.54 second, while the slower ones are
still on the way. By comparing with Figure \ref{exp2} (upper plot),
we can observe that slow UAVs are overtaken by faster ones.

\section{Conclusions}

The virtual tube passing problem, which includes passing a virtual
tube, inter-agent conflict avoidance and keeping within the virtual
tube, is studied in this paper. Based on the velocity control model
of UAVs with control saturation, practical distributed control is
proposed for multiple UAVs to pass a virtual tube. Every UAV has the
same and simple control protocol. Lyapunov-like functions are designed
elaborately, and formal analysis and proofs are made to show that
the virtual tube passing problem can be solved, namely passing the
virtual tube without getting trapped, avoiding conflict and keeping
within the virtual tube. Besides the functional requirement, the safety
requirement is also satisfied. By the proposed distributed control,
a UAV can keep away from another one or return back to the virtual
tube as soon as possible, once it enters into the safety area of another
UAV or has a conflict with the virtual tube accidentally during it
is passing the virtual tube. This is very necessary to guarantee the
flight safety. Simulations and experiments are given to show the advantages
of the proposed method over other algorithms in terms of calculation
speed of finding feasible solutions, and the effectiveness of the
proposed method from the functional requirement and the safety requirement.

\section{Appendix}

\subsection{Proof of Proposition 1}

\textit{Proof}. According to (\ref{positionmodel_ab_con_i}), we have
\[
\mathbf{v}_{i}^{\text{T}}\mathbf{\dot{v}}_{i}=-l_{i}\left(\mathbf{v}_{i}^{\text{T}}\mathbf{v}_{i}-\mathbf{v}_{i}^{\text{T}}\mathbf{v}_{\text{c},i}\right).
\]
Then 
\[
\frac{\text{d}\left\Vert \mathbf{v}_{i}\right\Vert }{\text{d}t}=-l_{i}\left\Vert \mathbf{v}_{i}\right\Vert +\frac{1}{\left\Vert \mathbf{v}_{i}\right\Vert }\mathbf{v}_{i}^{\text{T}}\mathbf{v}_{\text{c},i}
\]
whose solution is 
\begin{align*}
\left\Vert \mathbf{v}_{i}\left(t\right)\right\Vert  & =e^{-l_{i}t}\left\Vert \mathbf{v}_{i}\left(0\right)\right\Vert +
{\displaystyle \int\nolimits _{0}^{t}}
e^{-l_{i}\left(t-\tau\right)}\frac{1}{\left\Vert \mathbf{v}_{i}\left(\tau\right)\right\Vert }\mathbf{v}_{i}^{\text{T}}\mathbf{v}_{\text{c},i}\left(\tau\right)\text{d}\tau\\
 & \leq e^{-l_{i}t}\left\Vert \mathbf{v}_{i}\left(0\right)\right\Vert +
{\displaystyle \int\nolimits _{0}^{t}}
e^{-l_{i}\left(t-\tau\right)}{v_{\text{m},i}}\text{d}\tau.
\end{align*}
If $\left\Vert \mathbf{v}_{i}\left(0\right)\right\Vert \leq{v_{\text{m},i},}$
then 
\[
\left\Vert \mathbf{v}_{i}\left(t\right)\right\Vert \leq{v_{\text{m},i}.}
\]
With the result, one has 
\begin{align*}
\left\Vert \boldsymbol{\xi}_{i}\left(t\right)-\boldsymbol{\xi}_{j}\left(t\right)\right\Vert  & \leq\left\Vert \mathbf{p}_{i}\left(t\right)-\mathbf{p}_{j}\left(t\right)\right\Vert +\left\Vert \frac{1}{{l}_{i}}\mathbf{v}_{i}\right\Vert +\left\Vert \frac{1}{{l}_{j}}\mathbf{v}_{j}\right\Vert \\
 & \leq\left\Vert \mathbf{p}_{i}\left(t\right)-\mathbf{p}_{j}\left(t\right)\right\Vert +2r_{\text{v}}{.}
\end{align*}
Since $\left\Vert \boldsymbol{\xi}_{i}\left(t\right)-\boldsymbol{\xi}_{j}\left(t\right)\right\Vert \geq r+2r_{\text{v}},$
one further has 
\[
\left\Vert \mathbf{p}_{i}\left(t\right)-{{\mathbf{p}}_{j}}\left(t\right)\right\Vert +r_{\text{v}}>r+r_{\text{v}}.
\]
Then $\left\Vert \mathbf{p}_{i}\left(t\right)-{{\mathbf{p}}_{j}}\left(t\right)\right\Vert >r.$


\subsection{Proof of Lemma 2}

The reason why these VTOL UAVs are able to avoid conflict with each
other, which will be proved by contradiction. Without loss of generality,
assume that $\left\Vert \boldsymbol{\tilde{\xi}}{_{\text{m,}ij_{1}}}\left(t_{\text{o}}\right)\right\Vert =2r_{\text{s}}$
occurs at $t_{\text{o}}>0$ first, i.e., a conflict between the $i$th$\ $UAV
and the $j_{1}$th$\ $UAV happening. Then, $\left\Vert \boldsymbol{\tilde{\xi}}{_{\text{m,}ij}}\left(t_{\text{o}}\right)\right\Vert >2r_{\text{s}}$
for $j\neq j_{1}$. Consequently, $V_{\text{m},ij}{\left(t_{\text{o}}\right)\geq0}$
if $j\neq j_{1}.$ Since ${V}\left(0\right)>0$ and ${{\dot{V}}\left(t\right)}\leq0$,
the function ${V}$ satisfies ${V}\left(t_{\text{o}}\right)\leq{V}\left(0\right),$
$t\in\left[0,\infty\right)$. By the definition of ${{V},}$ we have
\[
V{_{\text{m,}ij_{1}}}\left(t_{\text{o}}\right)\leq{V}\left(0\right).
\]
According to (\ref{sata}), given any $\epsilon_{rs}>0,$ there exists
a $\epsilon_{\text{s}}>0,$ such that 
\[
s\left(1,\epsilon_{\text{s}}\right)=1-\epsilon_{rs}.
\]
Then, at time $t_{\text{o}},$ the denominator of $V_{\text{m},ij_{1}}$
defined in (\ref{Vmij}) is 
\begin{align}
 & \left(1+\epsilon_{\text{m}}\right)\left\Vert \boldsymbol{\tilde{\xi}}{_{\text{m,}ij_{1}}}\left(t_{\text{o}}\right)\right\Vert -2r_{\text{s}}s\left(\frac{\left\Vert \boldsymbol{\tilde{\xi}}{_{\text{m,}ij_{1}}}\left(t_{\text{o}}\right)\right\Vert }{2r_{\text{s}}},\epsilon_{\text{s}}\right)\nonumber \\
 & =2r_{\text{s}}\left(1+\epsilon_{\text{m}}\right)-2r_{\text{s}}\left(1-\epsilon_{rs}\right)\nonumber \\
 & =2r_{\text{s}}\left(\epsilon_{\text{m}}+\epsilon_{rs}\right)\label{bound1}
\end{align}
where $\epsilon_{rs}>0$ can be sufficiently small if $\epsilon_{\text{s}}$
is sufficiently small according to (\ref{sata}). According to the
definition in (\ref{Vmij}), we have 
\begin{equation}
\frac{1}{2r_{\text{s}}\left(\epsilon_{\text{m}}+\epsilon_{rs}\right)}=\frac{V{_{\text{m,}ij_{1}}}\left(t_{\text{o}}\right)}{k_{2}}\leq\frac{{V}\left(0\right)}{k_{2}}\label{fact}
\end{equation}
where $\sigma_{_{\text{m}}}\left(\left\Vert \boldsymbol{\tilde{\xi}}{_{\text{m,}ij_{1}}}\right\Vert \right)=1$
is used. Consequently, ${{V}\left(0\right)}$ is \emph{unbounded}
as $\epsilon_{\text{m}}\rightarrow0\ $and $\epsilon_{rs}\rightarrow0.$
On the other hand, for any $j$, we have $\left\Vert \boldsymbol{\tilde{\xi}}{_{\text{m,}ij}}\left(0\right)\right\Vert >2r_{\text{s}}$
by \textit{Assumption 3}. Let $\left\Vert \boldsymbol{\tilde{\xi}}{_{\text{m,}ij}}\left(0\right)\right\Vert =2r_{\text{s}}+{\varepsilon_{\text{m,}ij},}$
${\varepsilon_{\text{m,}ij}}>0$. Then, at time $t=0,$ the denominator
of $V_{\text{m},ij}$ defined in (\ref{Vmij}) is 
\begin{align*}
 & \left(1+\epsilon_{\text{m}}\right)\left\Vert \boldsymbol{\tilde{\xi}}{_{\text{m,}ij}}\left(0\right)\right\Vert -2r_{\text{s}}s\left(\frac{\left\Vert \boldsymbol{\tilde{\xi}}{_{\text{m,}ij}}\left(0\right)\right\Vert }{2r_{\text{s}}},\epsilon_{\text{s}}\right)\\
 & \geq\left(1+\epsilon_{\text{m}}\right)\left(2r_{\text{s}}+{\varepsilon_{\text{m,}ij}}\right)-2r_{\text{s}}\bar{s}\left(\frac{\left\Vert \boldsymbol{\tilde{\xi}}{_{\text{o,}ij}}\left(0\right)\right\Vert }{2r_{\text{s}}}\right)\\
 & =2r_{\text{s}}\epsilon_{\text{m}}+\left(1+\epsilon_{\text{m}}\right){\varepsilon_{\text{m,}ij}}.
\end{align*}
Then 
\[
V{_{\text{m,}ij}}\left(0\right)\leq\frac{k_{2}}{2r_{\text{s}}\epsilon_{\text{m}}+\left(1+\epsilon_{\text{m}}\right){\varepsilon_{\text{m,}ij}}}.
\]
Consequently, $V{_{\text{m,}ij}}\left(0\right)$ is still bounded
as $\epsilon_{\text{m}}\rightarrow0\ $no matter what $\epsilon_{rs}\ $is.
According to the definition of ${V}\left(0\right),$ ${V}\left(0\right)$
is still \emph{bounded} as $\epsilon_{\text{m}}\rightarrow0\ $and
$\epsilon_{rs}\rightarrow0.$ This is a contradiction. Thus 
\begin{equation}
\left\Vert \boldsymbol{\tilde{\xi}}{_{\text{m,}ij}}\left(t\right)\right\Vert >2r_{\text{s}},i\neq j\label{bounded1}
\end{equation}
for $i,j=1,2,\cdots,N,$ $t\in\left[0,\infty\right).$ Therefore,
the UAV can avoid another UAV by the velocity command (\ref{*}).

The reason why a UAV can stay within the virtual tube is similar to
the above proof. It can be proved by contradiction as well. Without
loss of generality, assume that $\left\Vert \boldsymbol{\tilde{\xi}}{_{\text{t},i}}\left(t_{\text{o}}^{\prime}\right)\right\Vert =r_{\text{t}}-{{r}_{\text{s}}}$
occurs at $t_{\text{1}}^{\prime}>0,$ i.e., a conflict happening first,
while $\left\Vert \boldsymbol{\tilde{\xi}}{_{\text{t},i}}\left(t_{\text{o}}^{\prime}\right)\right\Vert >r_{\text{t}}-{{r}_{\text{s}}}$
for $i=1,\cdots,M$. Similar to the above proof, one can also get
a contradiction. 

\begin{IEEEbiography}[{{{{{\includegraphics[clip,width=1in,height=1.25in]{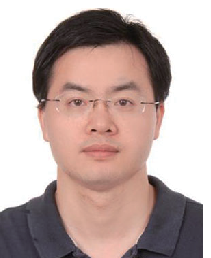}}}}}}]{Quan Quan}
received the B.S. and Ph.D. degrees in control science and engineering
from Beihang University, Beijing, China, in 2004 and 2010, respectively.
He has been an Associate Professor with Beihang University since 2013,
where he is currently with the School of Automation Science and Electrical
Engineering. His research interests include reliable flight control,
vision-based navigation, repetitive learning control, and timedelay
systems. 
\end{IEEEbiography}

\begin{IEEEbiography}[{{{{{\includegraphics[clip,width=1in,height=1.25in]{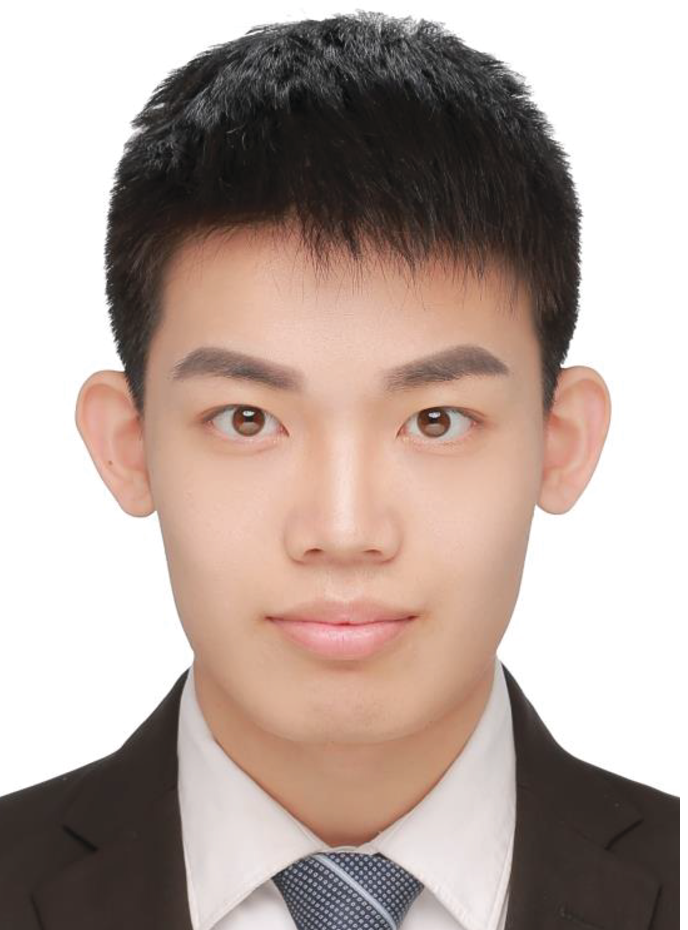}}}}}}]{Rao Fu}
is working toward to the Ph.D. degree at the School of Automation
Science and Electrical Engineering, Beihang University (formerly Beijing
University of Aeronautics and Astronautics), Beijing, China. His main
research interests include UAV traffic control and swarm. 
\end{IEEEbiography}

\begin{IEEEbiography}[{{{{{\includegraphics[clip,width=1in,height=1.25in]{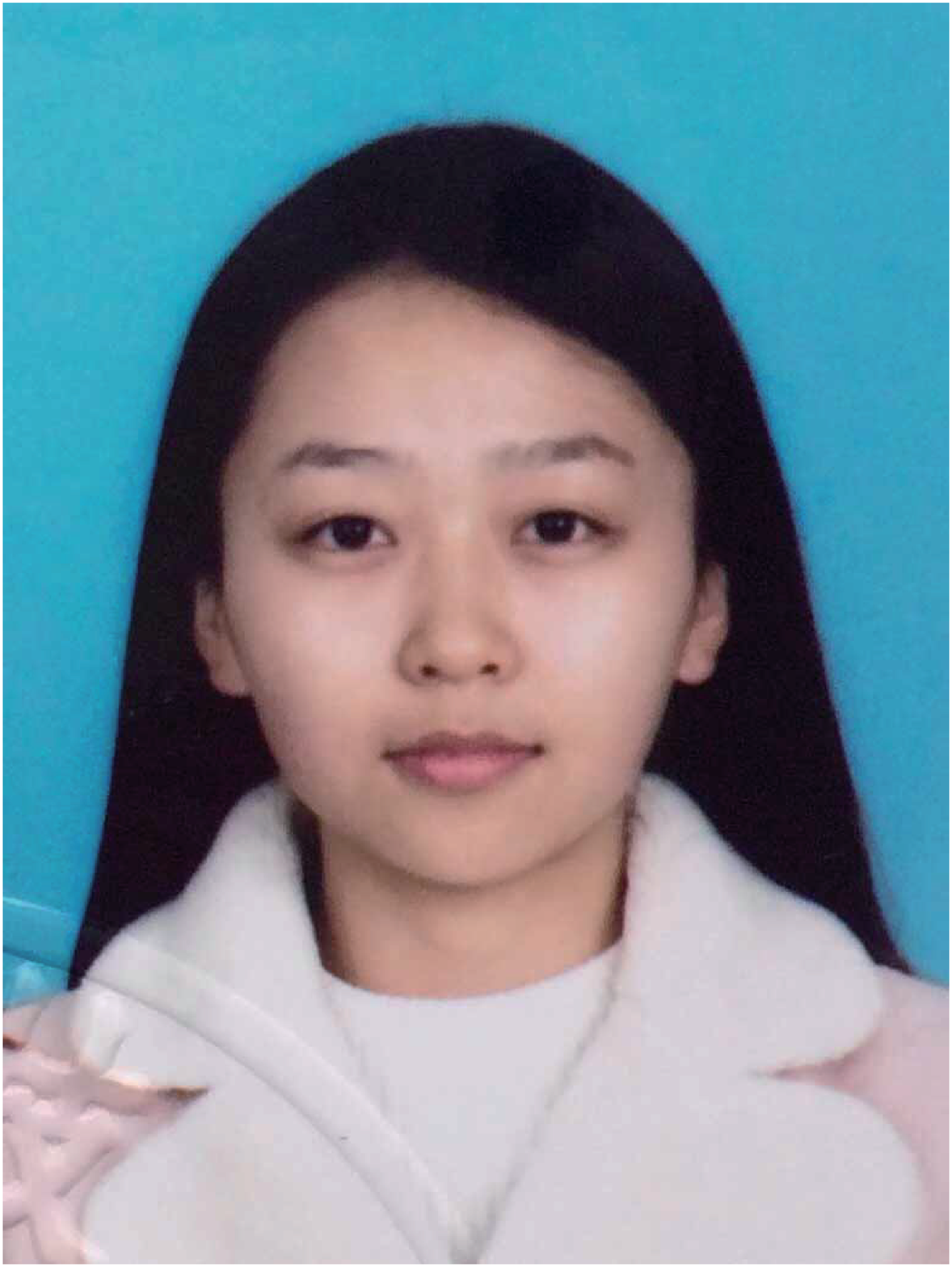}}}}}}]{Mengxin Li}
is working toward to the M.S. degree at the School of Automation
Science and Electrical Engineering, Beihang University (formerly Beijing
University of Aeronautics and Astronautics), Beijing, China. Her main
research interests include flight safety and control of multicopter. 
\end{IEEEbiography}

\begin{IEEEbiography}[{{{{\includegraphics[clip,width=1in,height=1.25in]{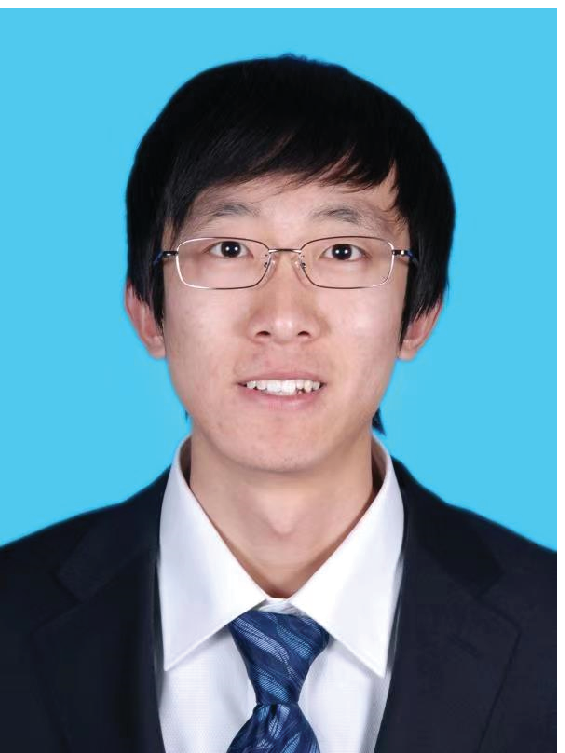}}}}}]{Donghui Wei}
recived the B.S. and Ph.D. degrees in Aircraft Design from Beihang
University, Beijing, China, in 2009 and 2020, respectively. He has
been a senior engineer in Science and Technology on Complex System
Control and Intelligent Agent Cooperation Laboratory. His research
interests include ultra-low altitude flight control and swarm flight
control. 
\end{IEEEbiography}

\begin{IEEEbiography}[{{{{{\includegraphics[clip,width=1in,height=1.25in]{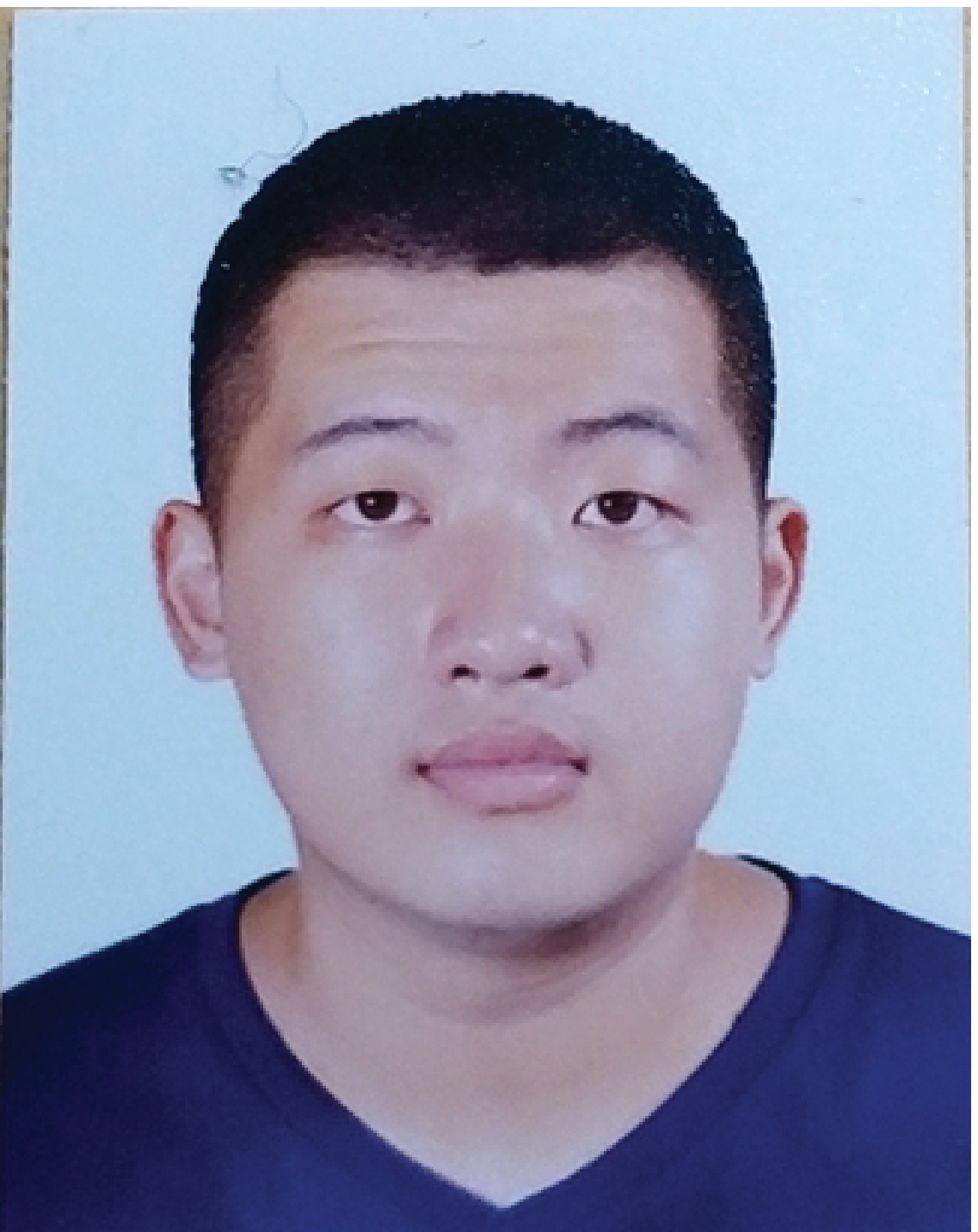}}}}}}]{Yan Gao}
is working toward to the Ph.D. degree at the School of Automation
Science and Electrical Engineering, Beihang University (formerly Beijing
University of Aeronautics and Astronautics), Beijing, China. His main
research interests include UAVs swarm and quadcopter control. 
\end{IEEEbiography}

\begin{IEEEbiography}[{{{{{\includegraphics[clip,width=1in,height=1.25in]{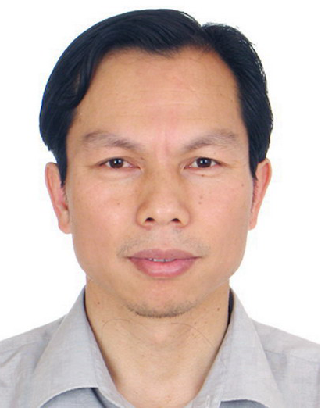}}}}}}]{Kai-Yuan Cai}
received the B.S., M.S., and Ph.D. degrees in control science and
engineering from Beihang University, Beijing, China, in 1984, 1987,
and 1991, respectively. He has been a Full Professor at Beihang University
since 1995. He is a Cheung Kong Scholar (Chair Professor), jointly
appointed by the Ministry of Education of China and the Li Ka Shing
Foundation of Hong Kong in 1999. His main research interests include
software testing, software reliability, reliable flight control, and
software cybernetics. 
\end{IEEEbiography}

\end{document}